\documentclass[times,twocolumn,final,authoryear]{elsarticle}

\usepackage{prletters}
\usepackage{framed,multirow}

\usepackage[latin1]{inputenc}
\usepackage{graphicx}
\usepackage[english]{babel}
\usepackage{epsfig}

\usepackage{array}

\usepackage{amssymb}
\usepackage{latexsym}

\usepackage{url}
\usepackage{xcolor}
\definecolor{newcolor}{rgb}{.8,.349,.1}

\journal{Pattern Recognition Letters}

\begin{document}

\clearpage
\thispagestyle{empty}

\ifpreprint
  \vspace*{-1pc}
\else
\fi

\begin{table*}[!t]

\section*{Research Highlights (Required)}

To create your highlights, please type the highlights against each
\verb+\item+ command. 

\vskip1pc

\fboxsep=6pt
\fbox{
\begin{minipage}{.95\textwidth}
\vskip1pc

\begin{itemize}
 \item We present a novel method for human activity recognition based on inertal sensors
 \item Recurrence Plots are computed from sensor data and used with computer vision approaches
 \item Sensor data recognition is then transformed into texture recognition
 \item The proposed method outperformed the baselines for classifying 12 human activities
\end{itemize}
\vskip1pc
\end{minipage}
}

\end{table*}

\clearpage

\ifpreprint
  \setcounter{page}{1}
\else
  \setcounter{page}{1}
\fi

\begin{frontmatter}

\title{Human activity recognition from mobile inertial sensors using recurrence plots}

\author[1]{Ot\'{a}vio A. B. \snm{Penatti}\corref{cor1}} 
\cortext[cor1]{Corresponding author: 
  Tel.: +55-19-98351-4422;  
  }
\ead{o.penatti@samsung.com}
\author[1]{Milton F. S. \snm{Santos}}

\address[1]{Advanced Technologies, SAMSUNG Research Institute, Campinas, SP, 13097-160, Brazil}

\received{3 Sep 2017}
\finalform{XX Month 2017}
\accepted{XX Month 2017}
\availableonline{YY Month 2017}
\communicated{S. Sarkar}

\begin{abstract}
Inertial sensors are present in most mobile devices nowadays and such devices are used by people during most of their daily activities. 
In this paper, we present an approach for human activity recognition based on inertial sensors by employing recurrence plots (RP) and visual descriptors. 
The pipeline of the proposed approach is the following: compute RPs from sensor data, compute visual features from RPs and use them in a machine learning protocol. 
As RPs generate texture visual patterns, we transform the problem of sensor data classification to a problem of texture classification. 
Experiments for classifying human activities based on accelerometer data showed that the proposed approach obtains the highest accuracies, outperforming time- and frequency-domain features directly extracted from sensor data. 
The best results are obtained when using RGB RPs, in which each RGB channel corresponds to the RP of an independent accelerometer axis. 
\end{abstract}

\begin{keyword}
\MSC 41A05\sep 41A10\sep 65D05\sep 65D17
\KWD Keyword1\sep Keyword2\sep Keyword3

human activity recognition\sep 
inertial sensors\sep 
recurrence plots\sep 
texture recognition\sep 

\end{keyword}

\end{frontmatter}

\section{Introduction}

Human Activity Recognition (HAR)~\citep{LabradorHARBook2013} has been a hot research topic for many years.
The two most common modalities of HAR are the ones based on sensor and camera data.
In Sensor-based HAR (SB-HAR)~\citep{ShoaibSurvey2015}, systems must be able to recognize human activities using only information from inertial sensors, like accelerometers and gyroscopes.
In Vision-based HAR (VB-HAR)~\citep{PoppeSurvey2010}, videos and images are the main source of information and computer vision is employed for decision making.
Given the recent growth in popularity of smart phones and wearable devices, which are equipped with inertial sensors, SB-HAR is specially important considering the potential to provide new ways of interaction, new applications and new possibilities to understand human behavior, as people hold their mobile devices for most of the day.

The most common approaches employed in SB-HAR are based on extracting time- and/or frequency-domain features from sensor data and then using such features with machine learning algorithms.
Statistical measures are commonly considered as time-domain features, like mean, standard deviation, root mean square, histograms, etc.
Frequency-domain features are based on the Fourier transform.
Time-domain features are mostly recommended for real-time systems, as they tend to be faster to compute~\citep{ShoaibSurvey2015}.

In general, inertial sensor data, like data from accelerometers and gyroscopes, can be seen as time-series data, 
and a popular approach for representing this kind of data is by using Recurrence Plots.
Recurrence Plots (RPs)~\citep{EckmannRP1987} were introduced in 1987 as a tool for representing time series of dynamic systems as images.
In practice, RPs help the visual understanding of recurrence properties of data.
RPs are images with texture patterns, as we can see in the examples shown in Figure~\ref{fig_examples_rps}. 

\newcommand{\wRPs}{0.065\textwidth}
\begin{figure}[t]
    \begin{tiny}
    \begin{center}
        \begin{tabular}{@{}c@{ }c@{ }c@{ }c@{ }c@{ }c@{ }c@{}}
        
            brush teeth &
            climb stairs &
            comb hair &
            descend stairs &
            drink glass &
            eat meat &
            eat soup \\
            
          \includegraphics[width=\wRPs]{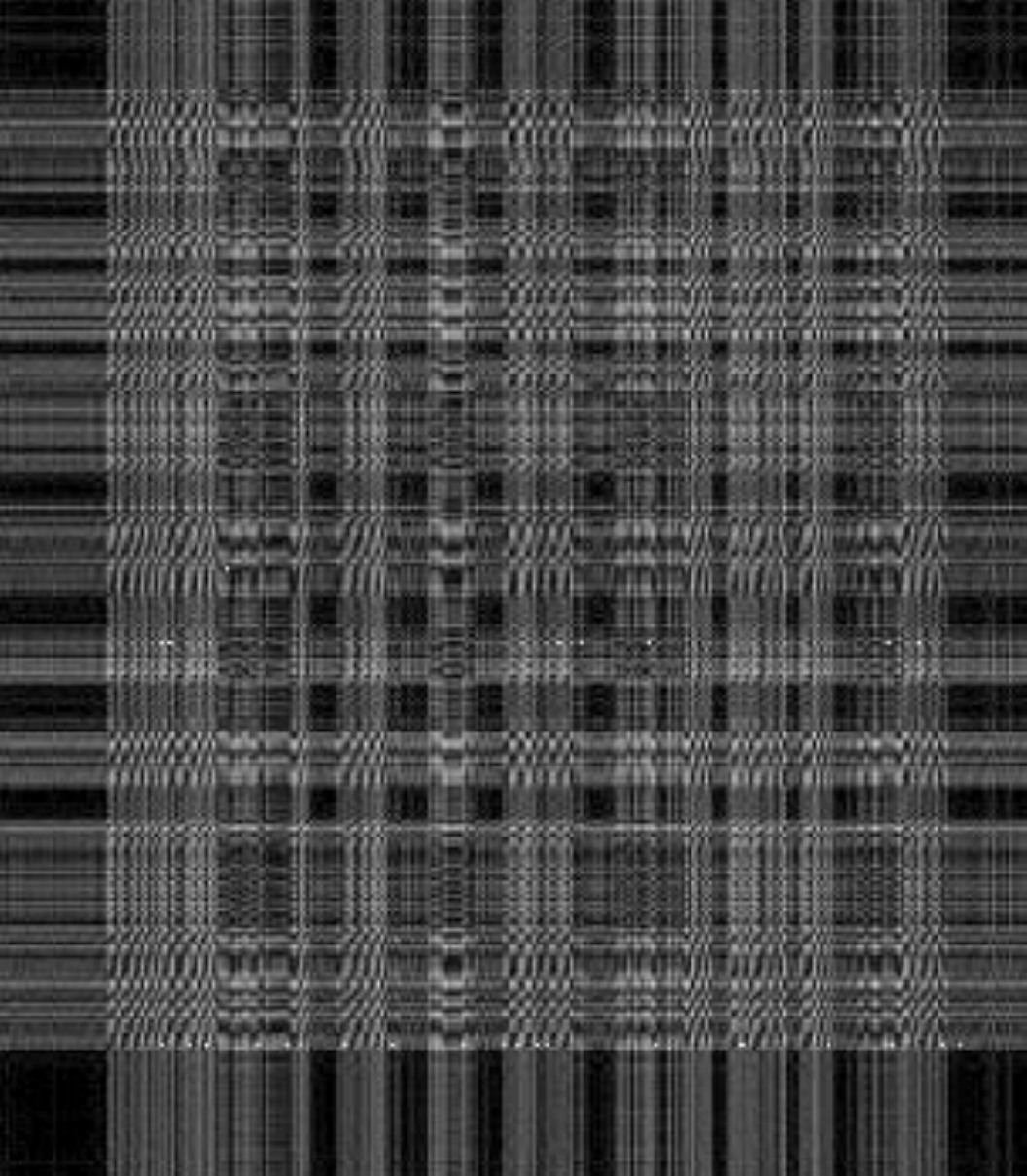} &
          \includegraphics[width=\wRPs]{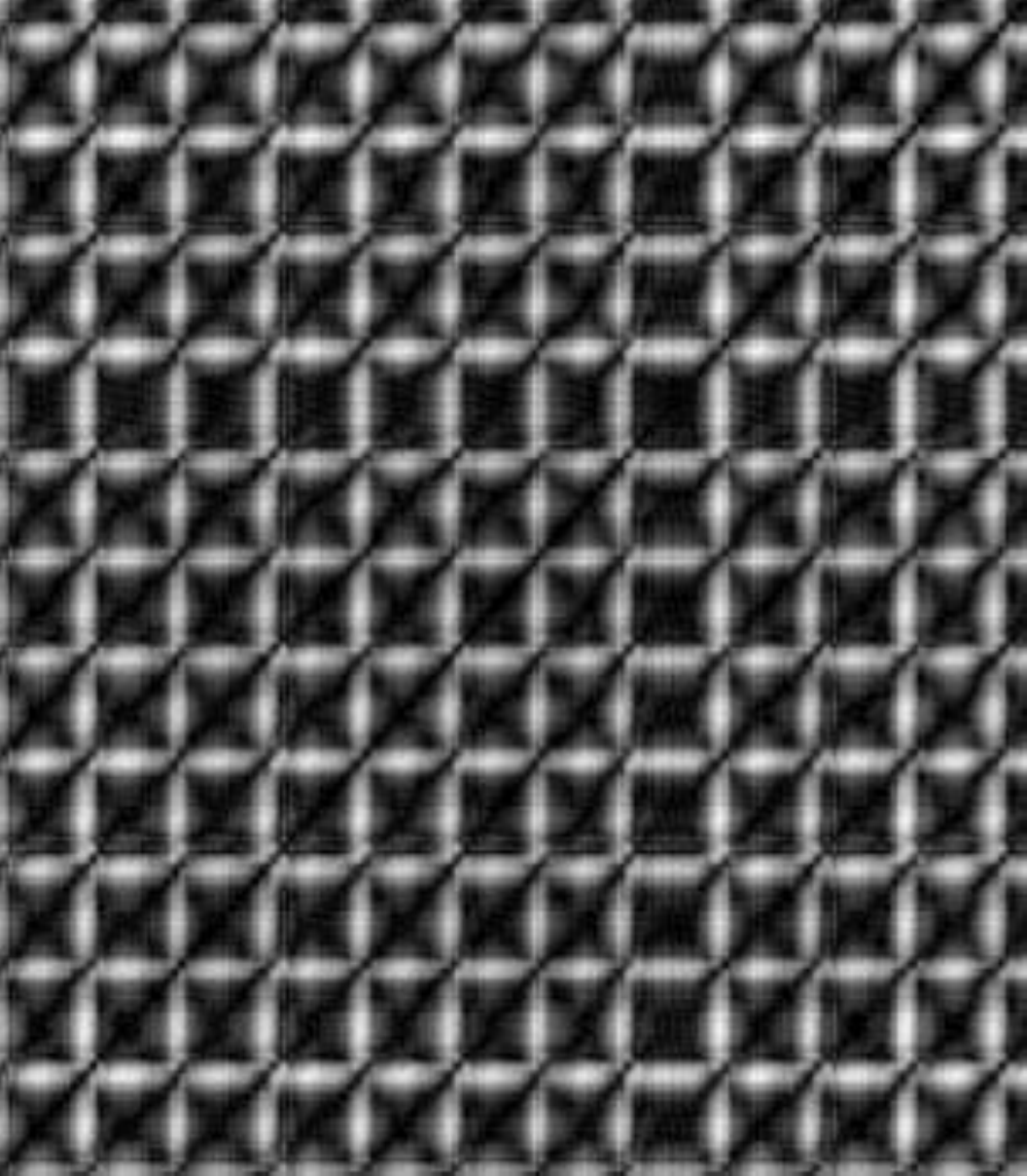} & 
          \includegraphics[width=\wRPs]{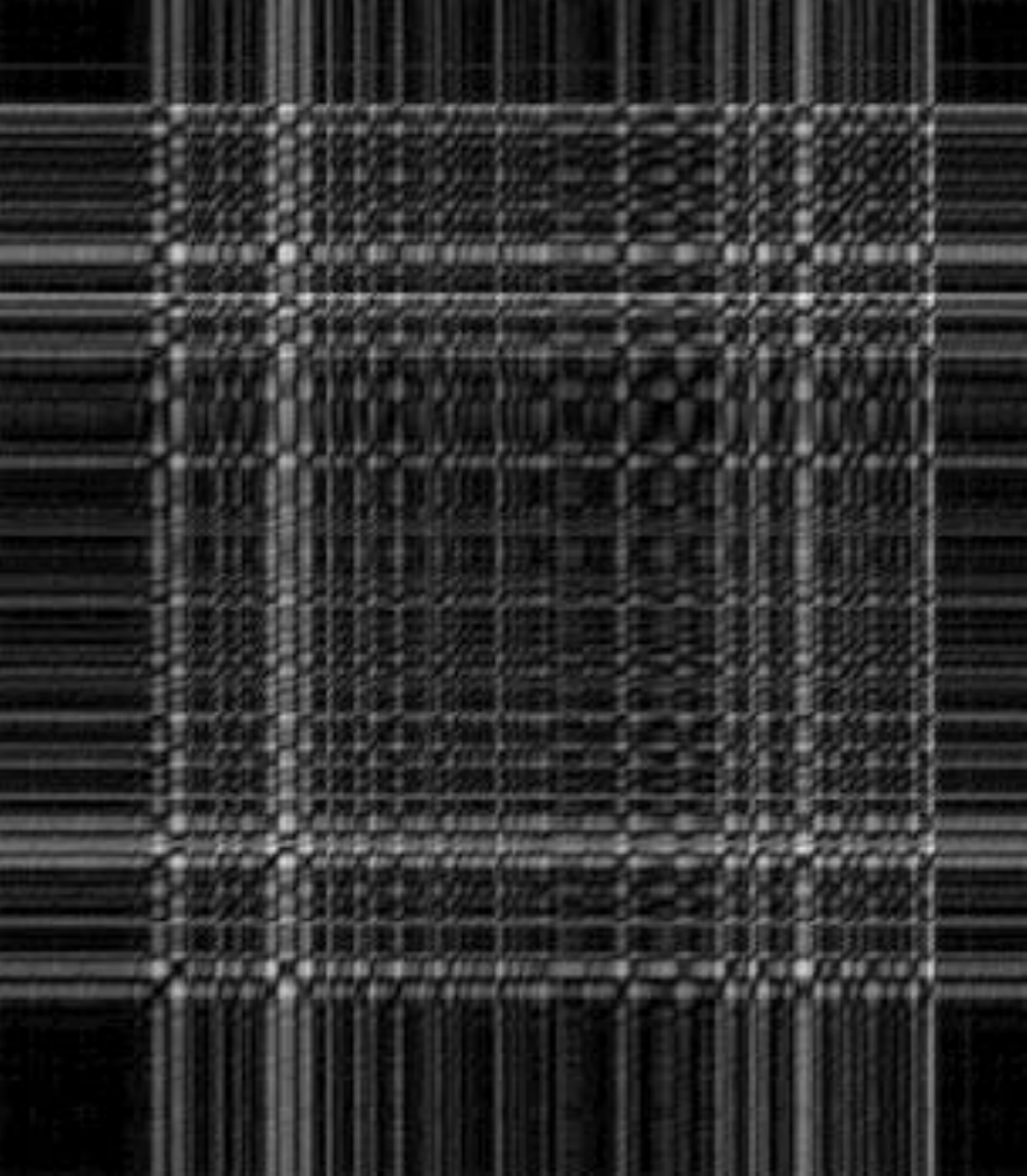} &
          \includegraphics[width=\wRPs]{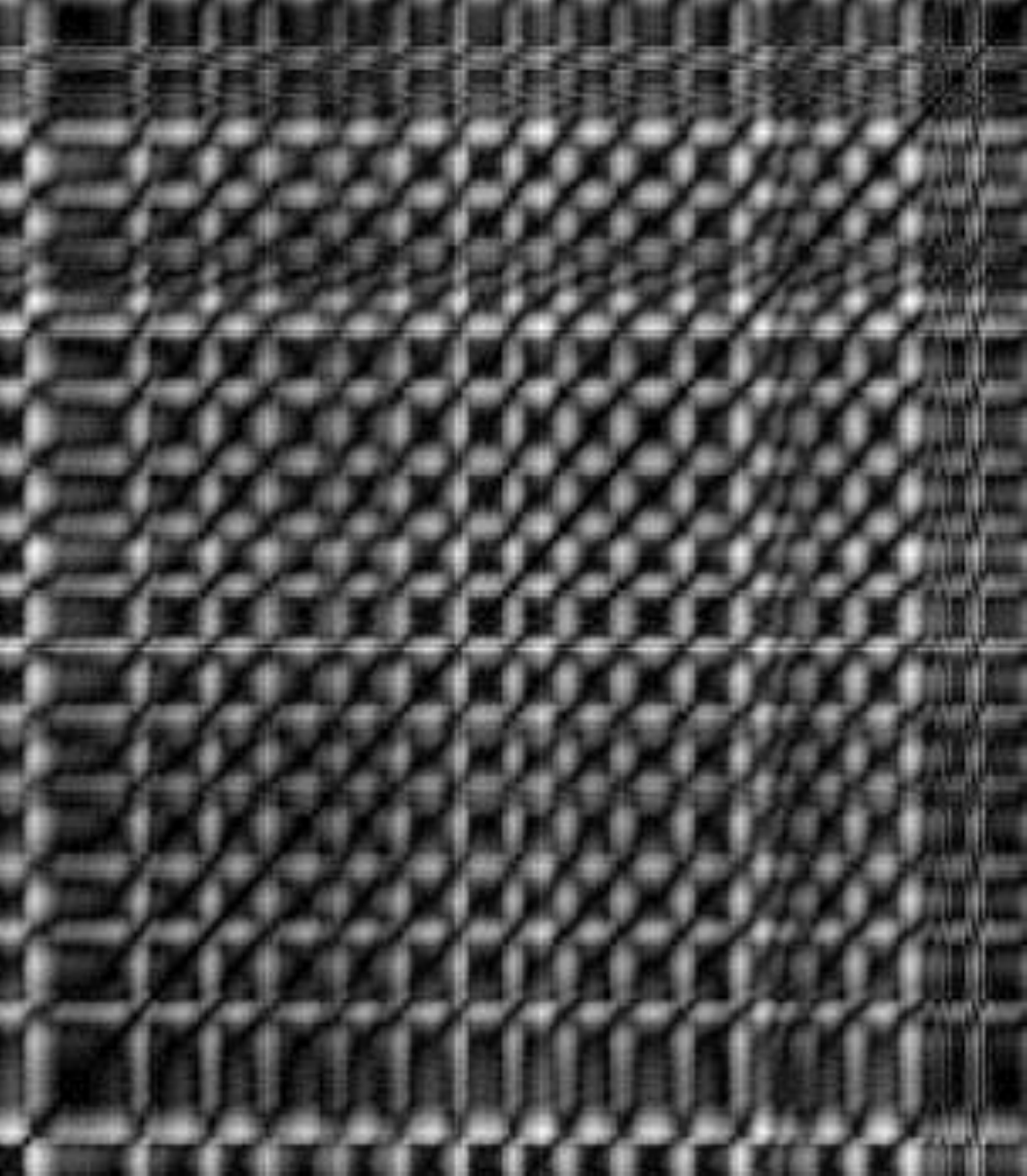} &
          \includegraphics[width=\wRPs]{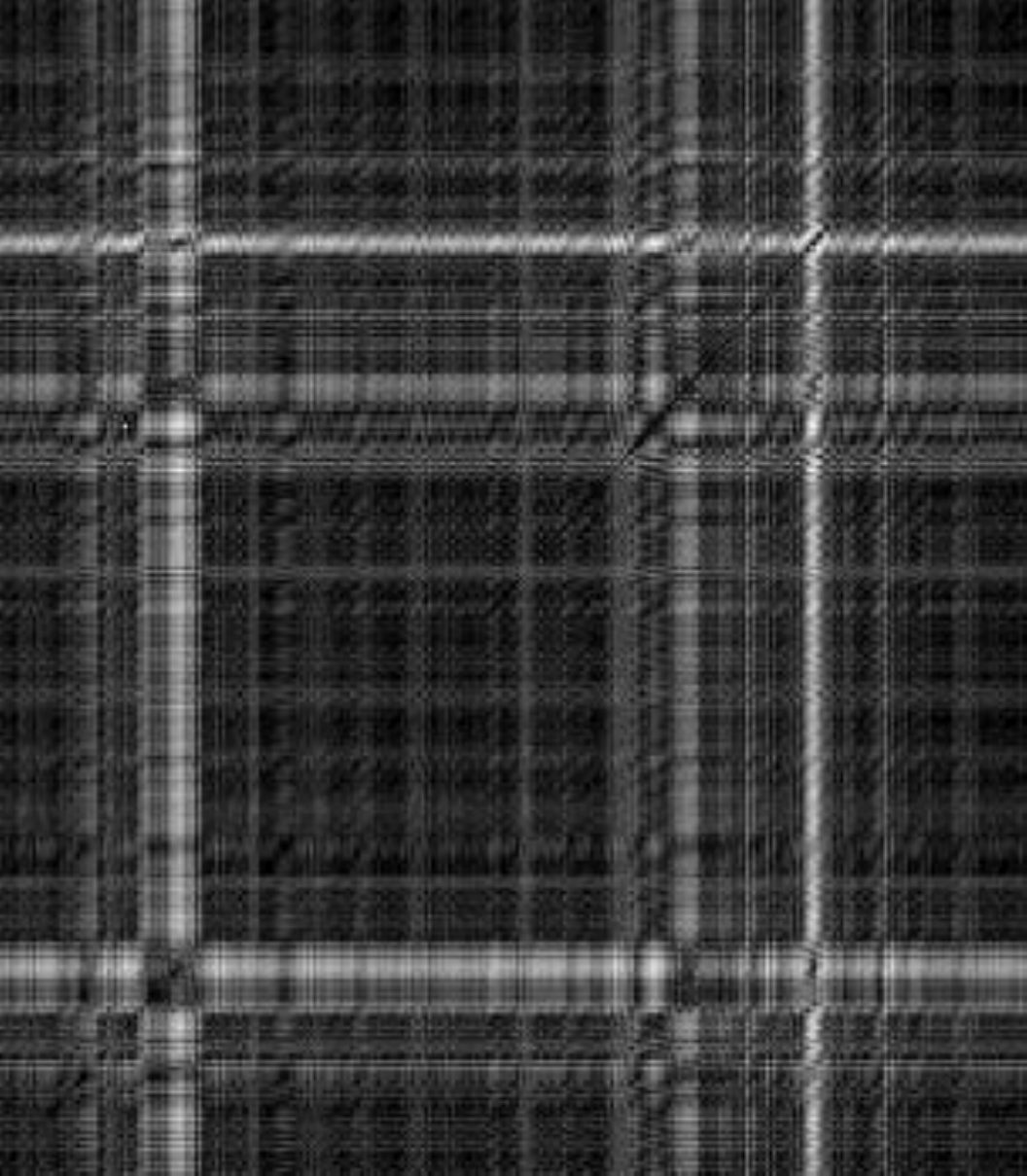} &
          \includegraphics[width=\wRPs]{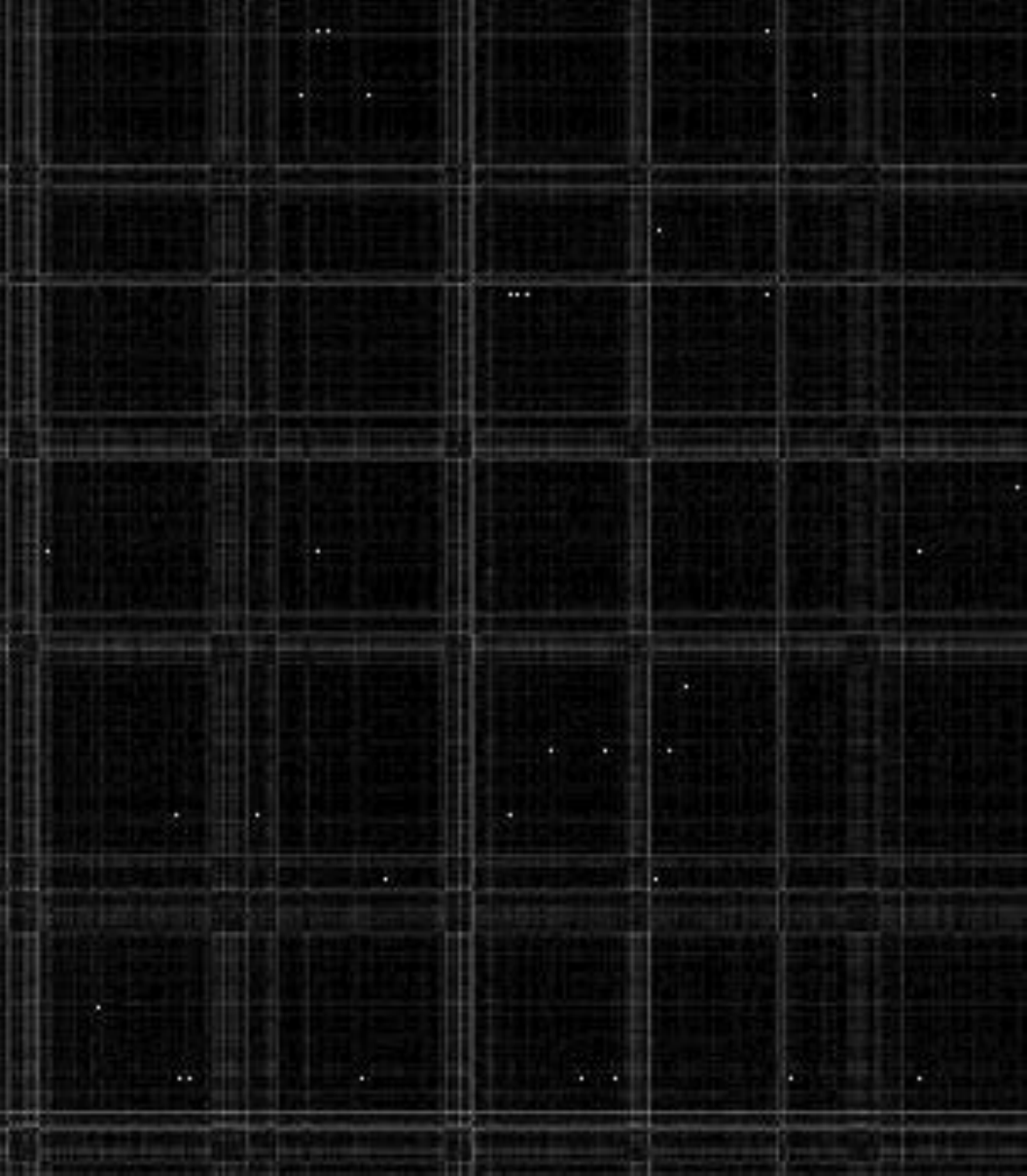} &
          \includegraphics[width=\wRPs]{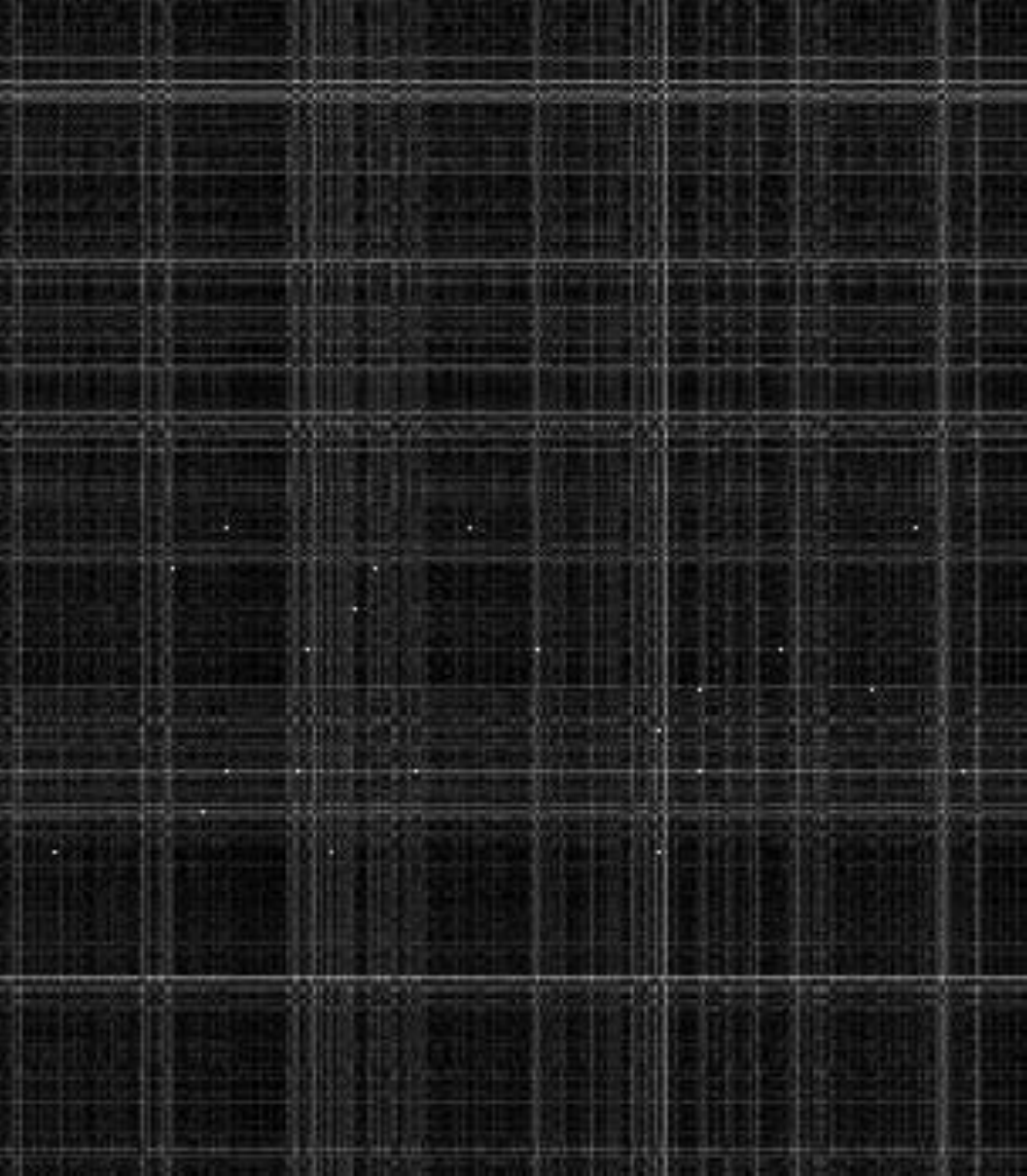} \\

          \includegraphics[width=\wRPs]{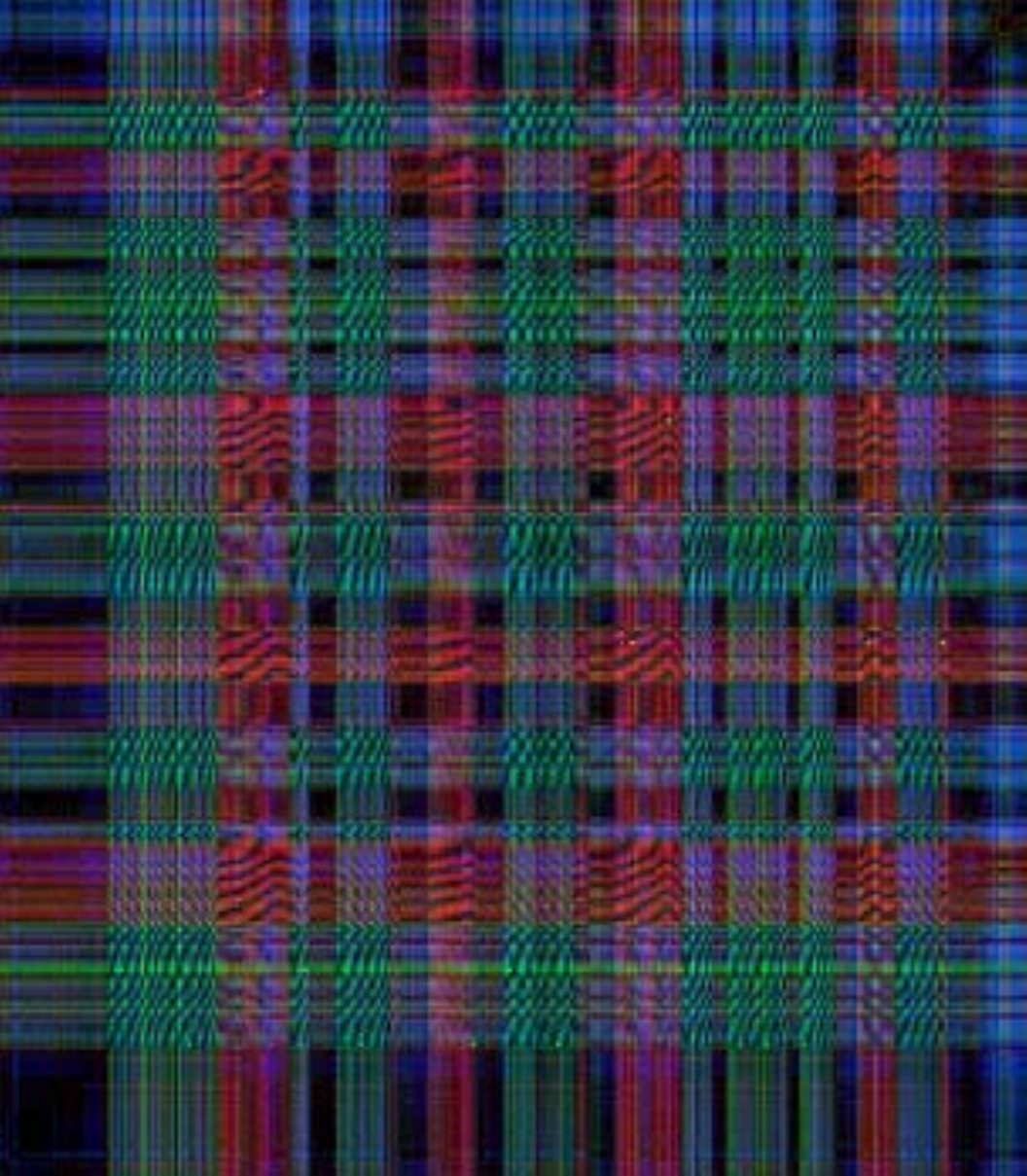} &
          \includegraphics[width=\wRPs]{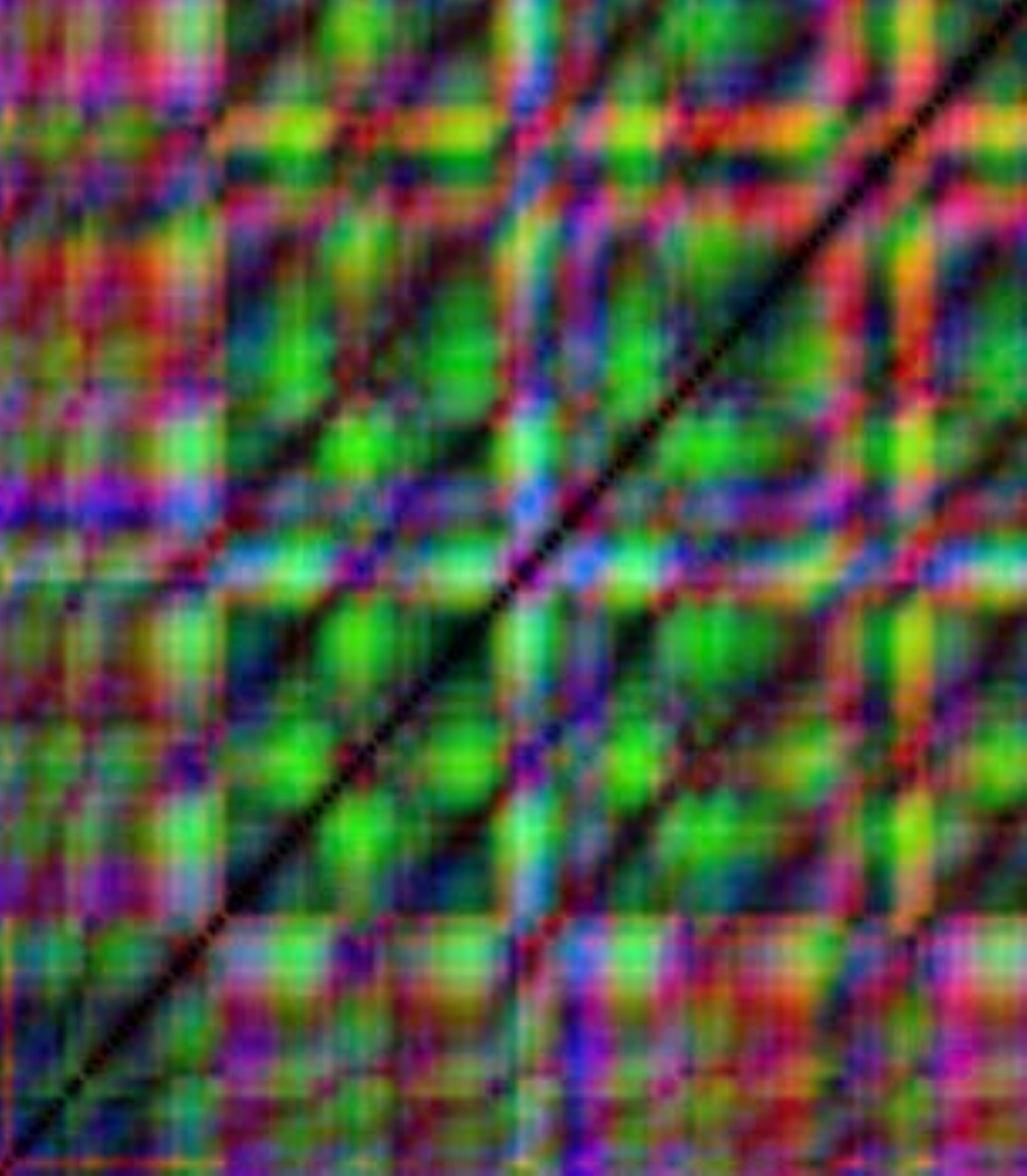} & 
          \includegraphics[width=\wRPs]{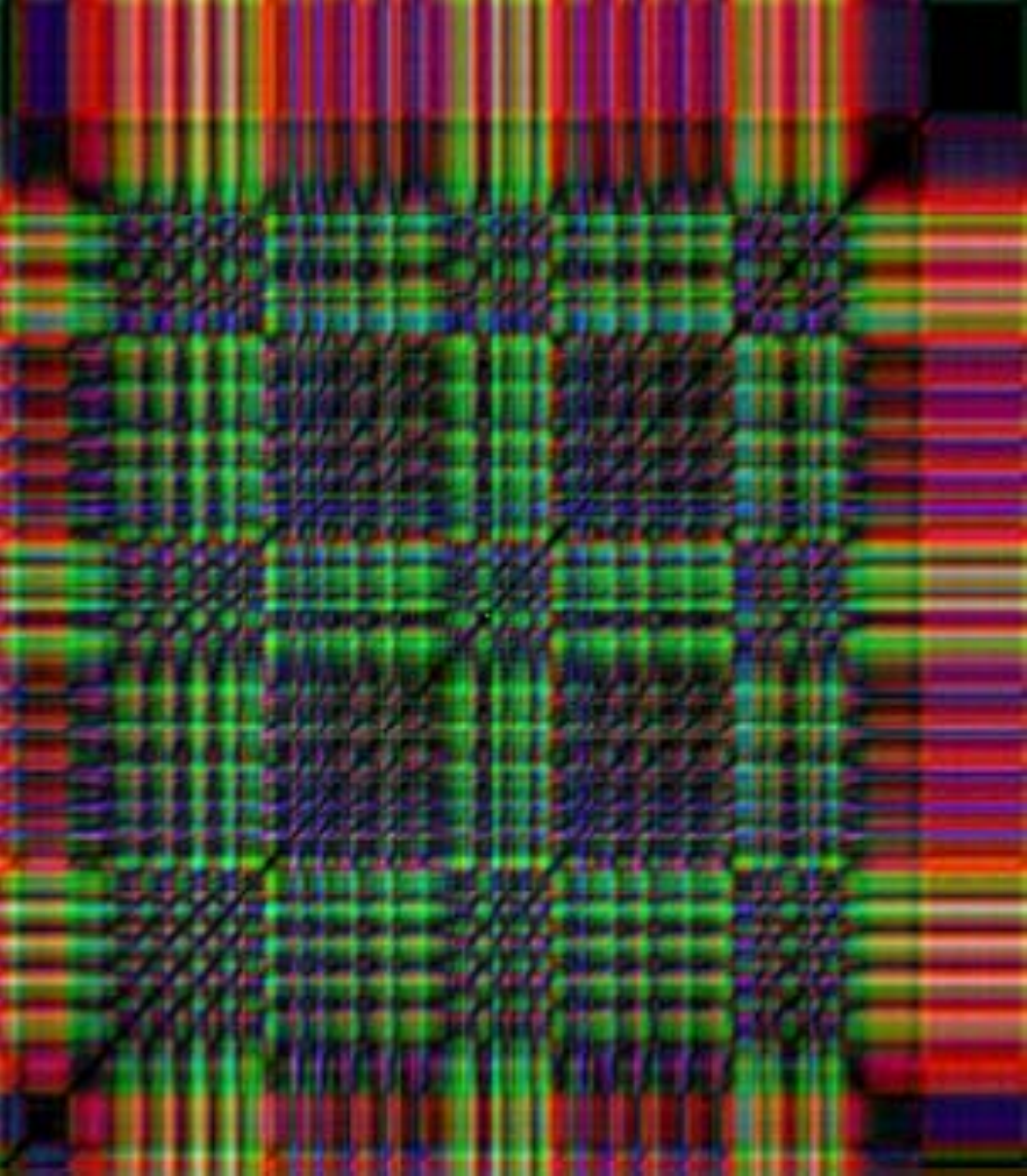} &
          \includegraphics[width=\wRPs]{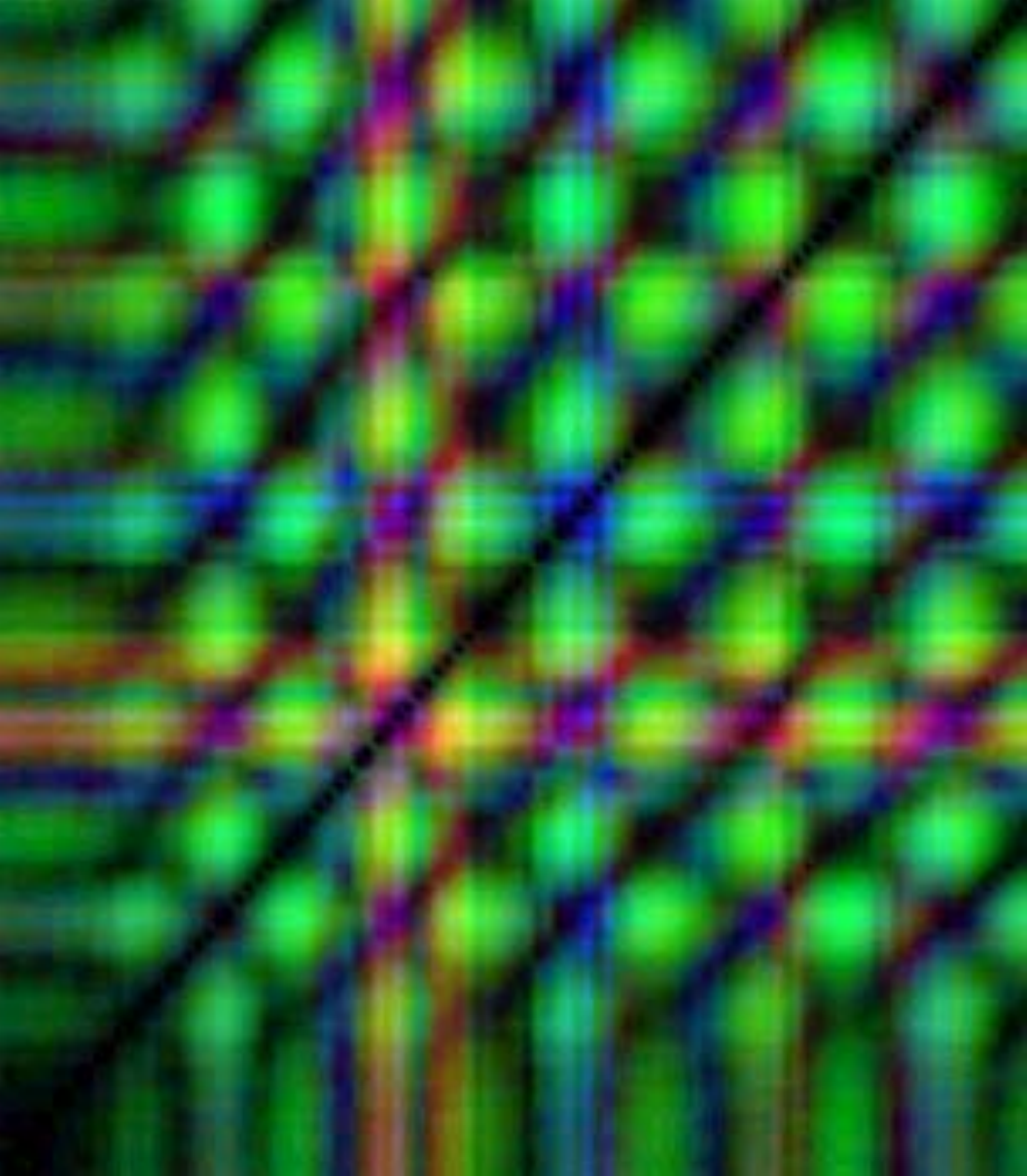} &
          \includegraphics[width=\wRPs]{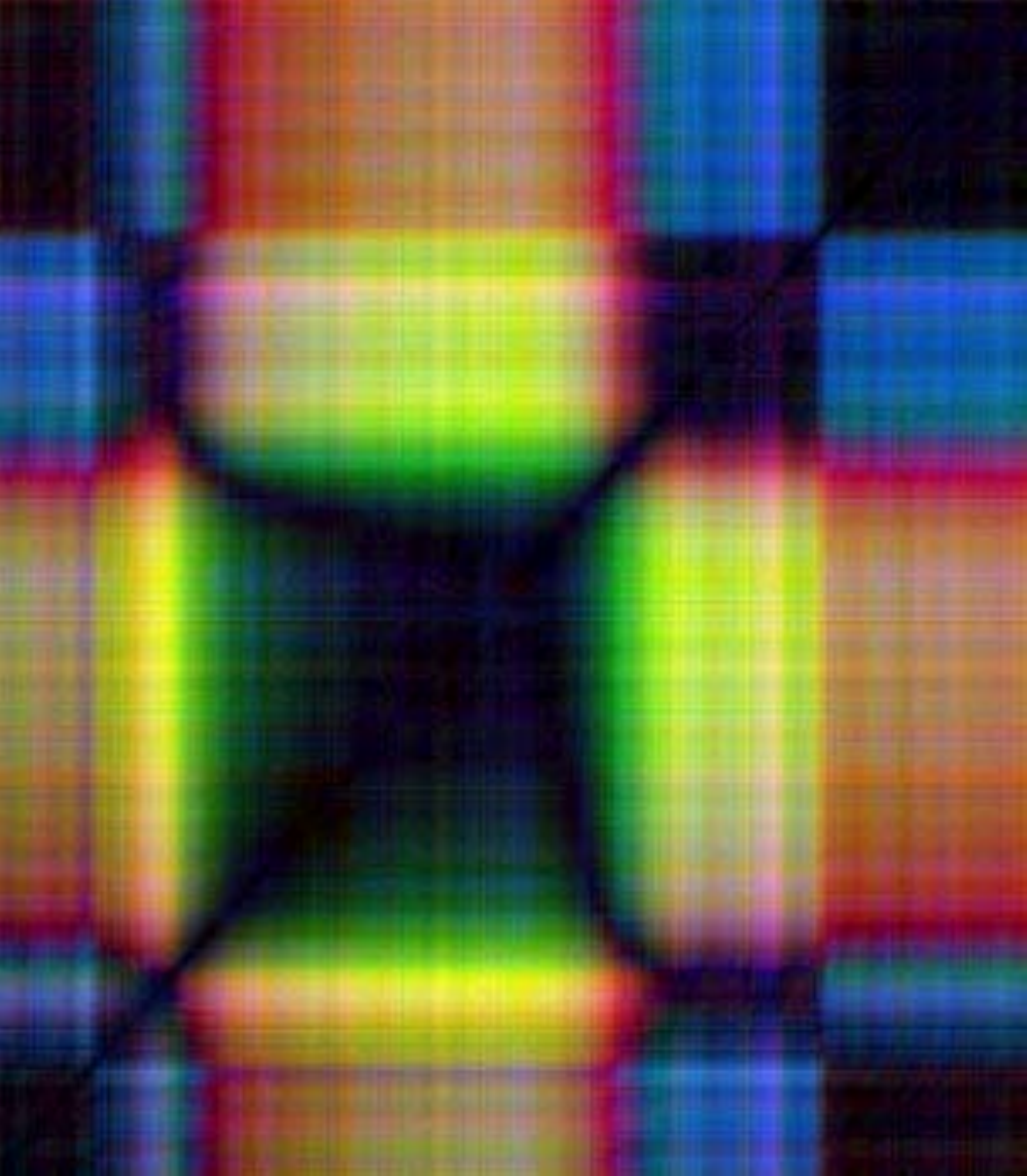} &
          \includegraphics[width=\wRPs]{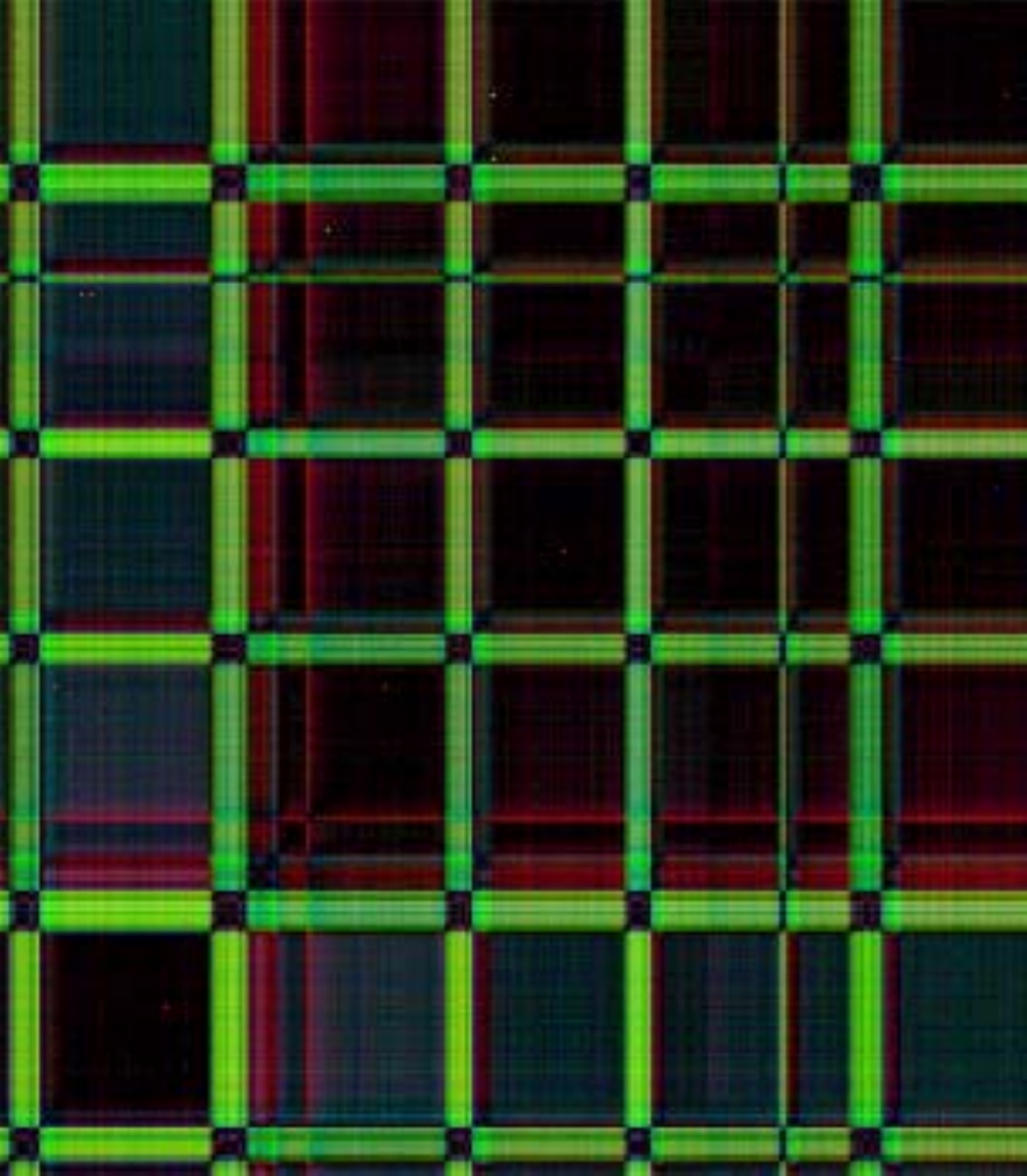} &  \vspace{0.5cm}
          \includegraphics[width=\wRPs]{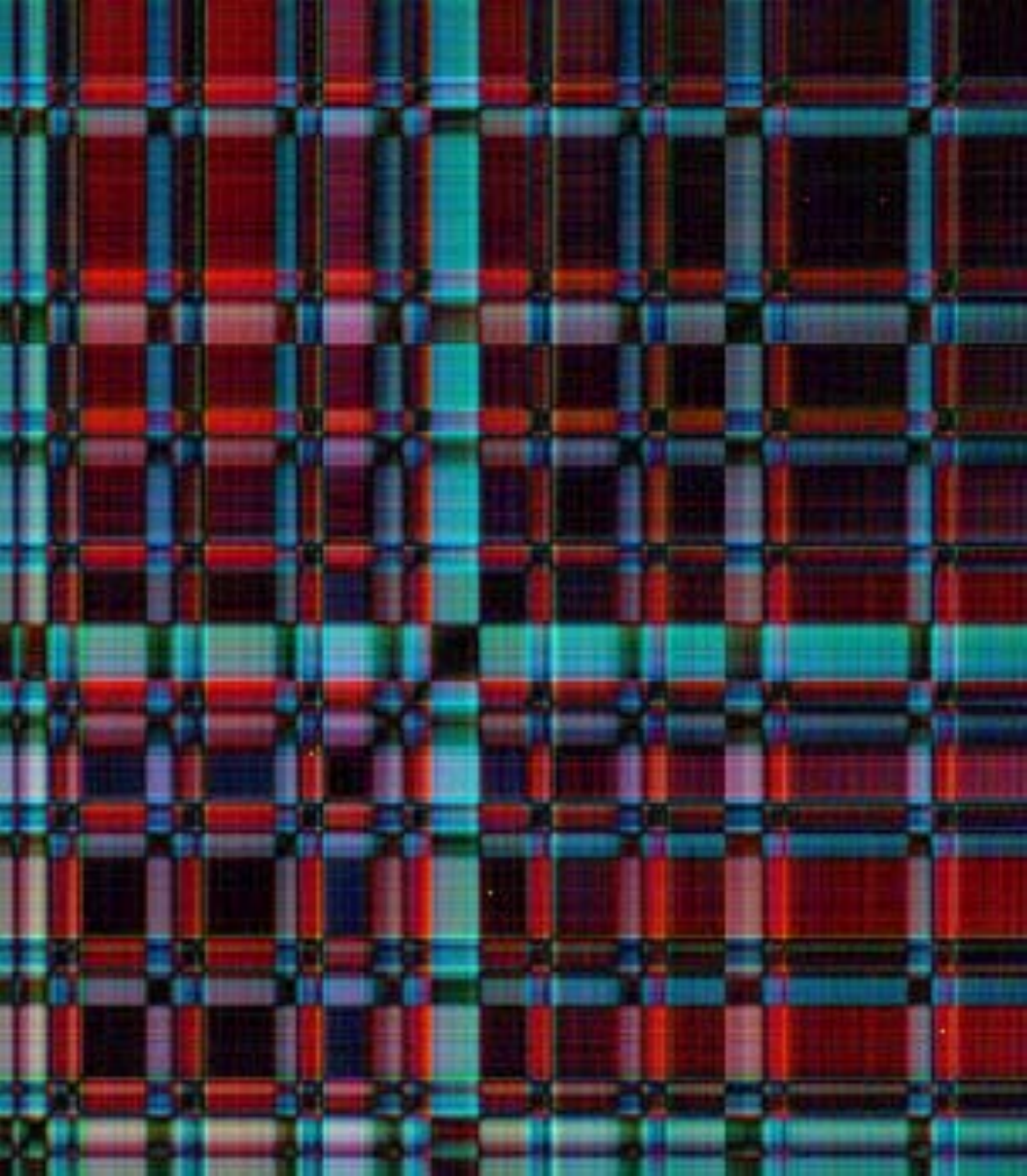} \\  

            get up bed &
            lie down bed &
            pour water &
            sit down chair &
            stand up chair &
            use telephone &
            walk \\  
            
          \includegraphics[width=\wRPs]{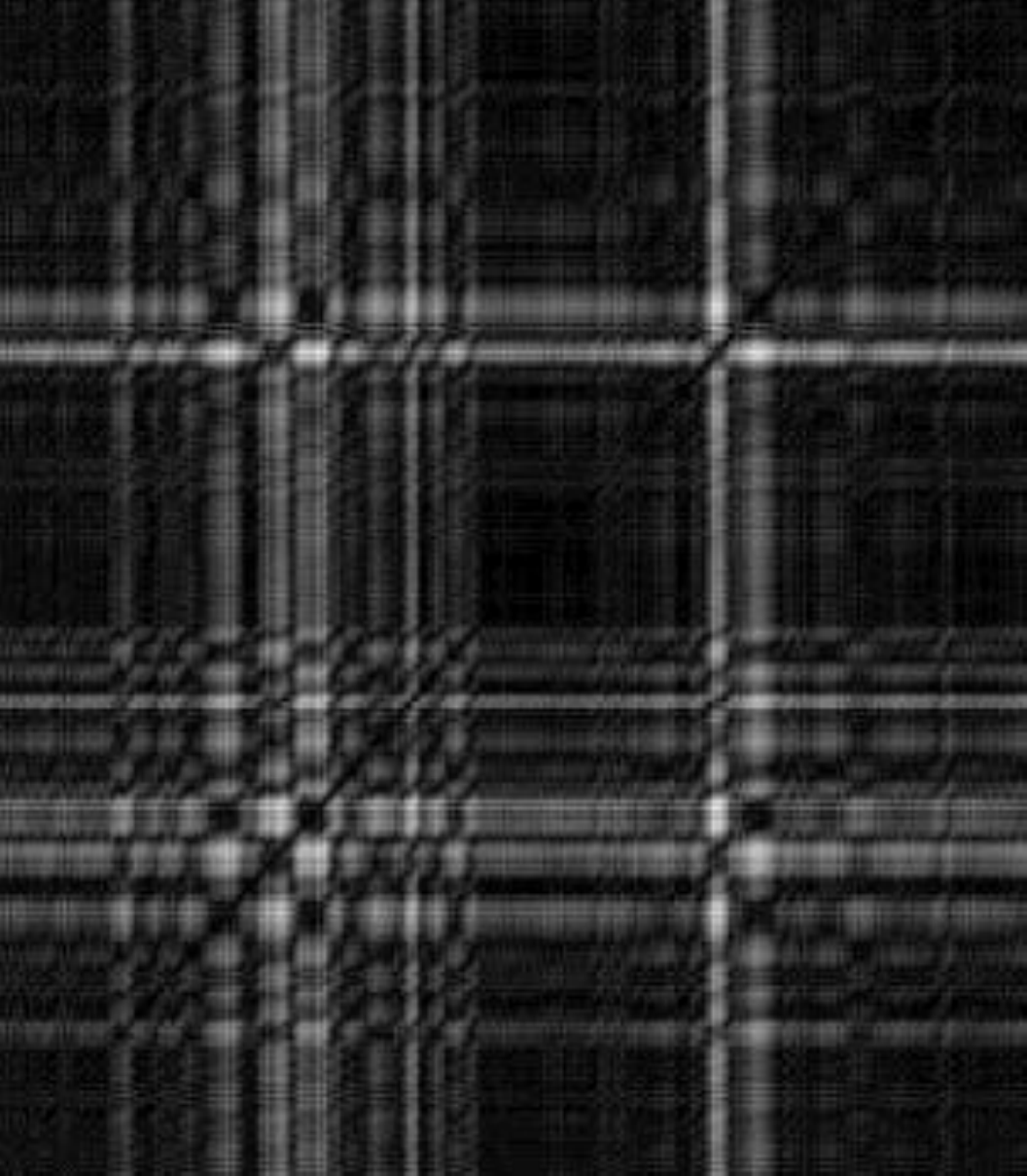} &
          \includegraphics[width=\wRPs]{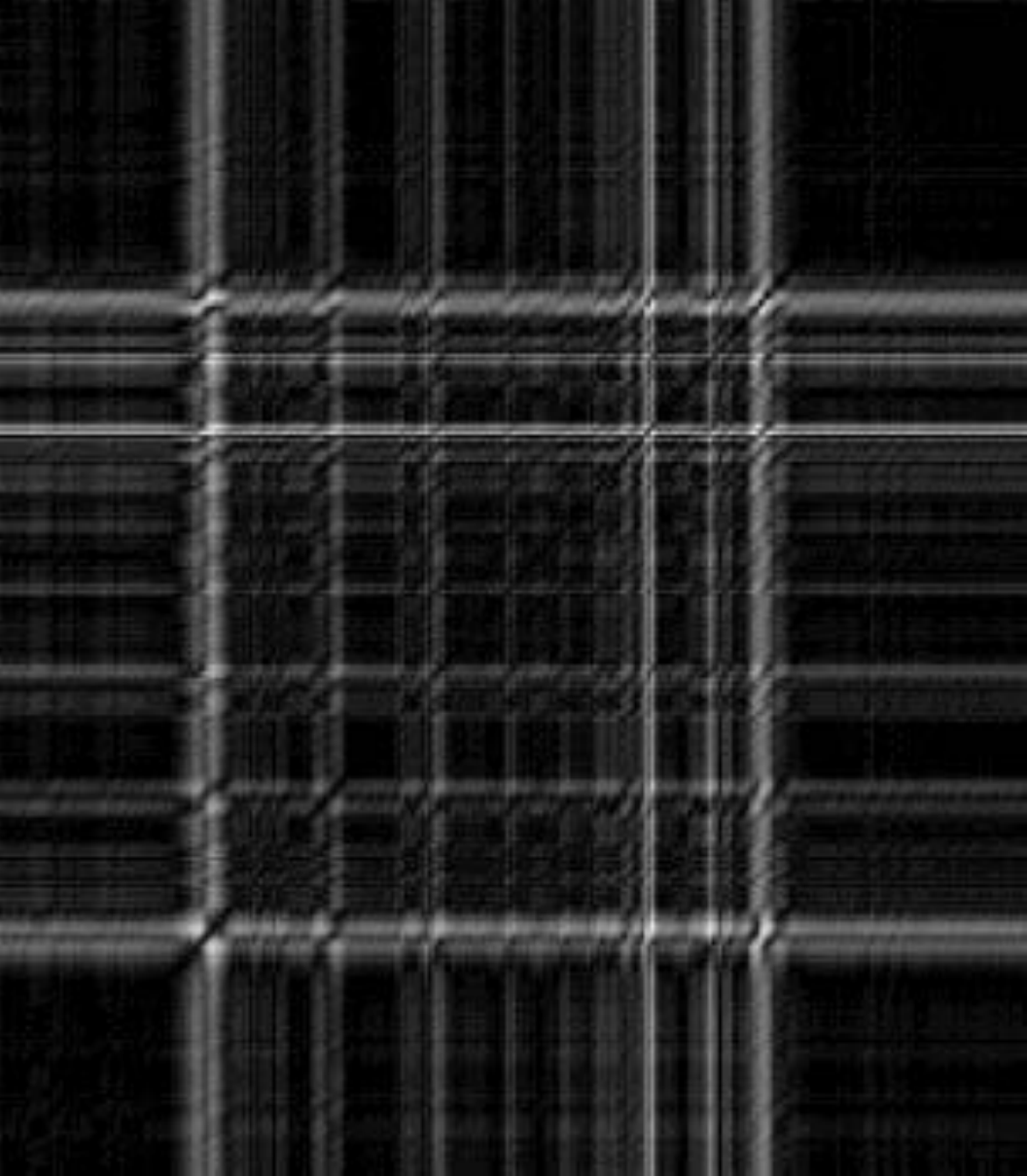} &
          \includegraphics[width=\wRPs]{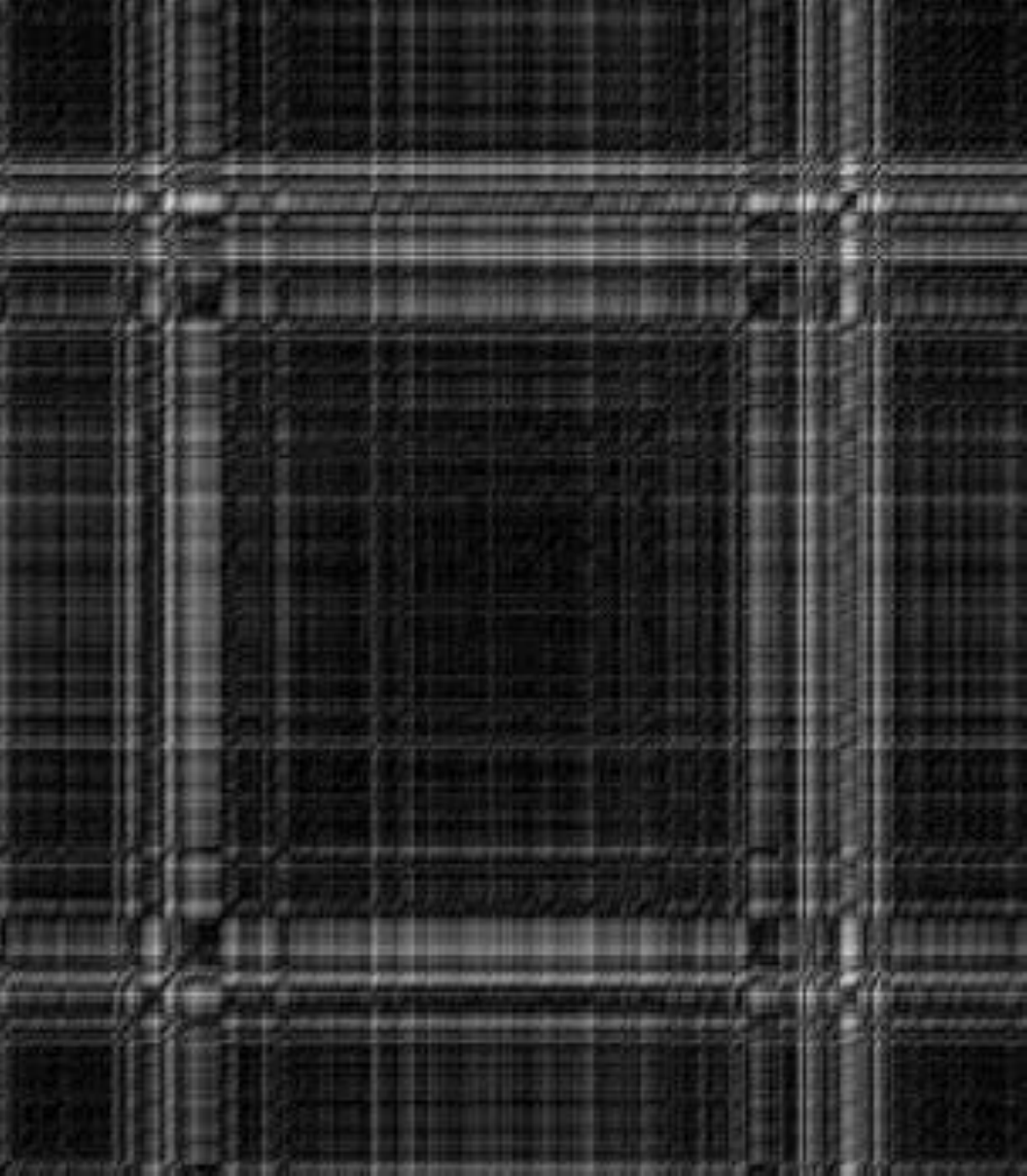} &
          \includegraphics[width=\wRPs]{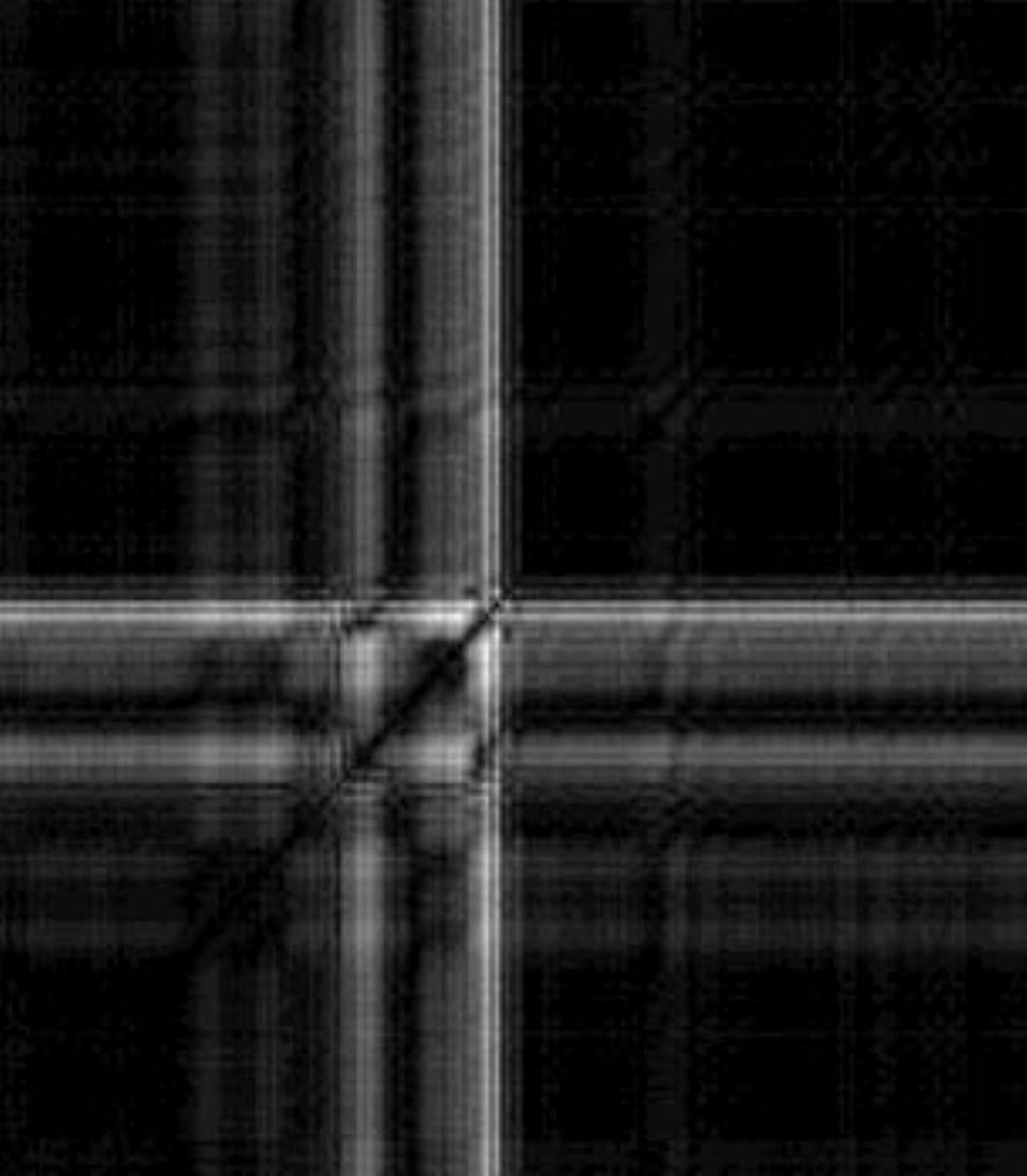} &
          \includegraphics[width=\wRPs]{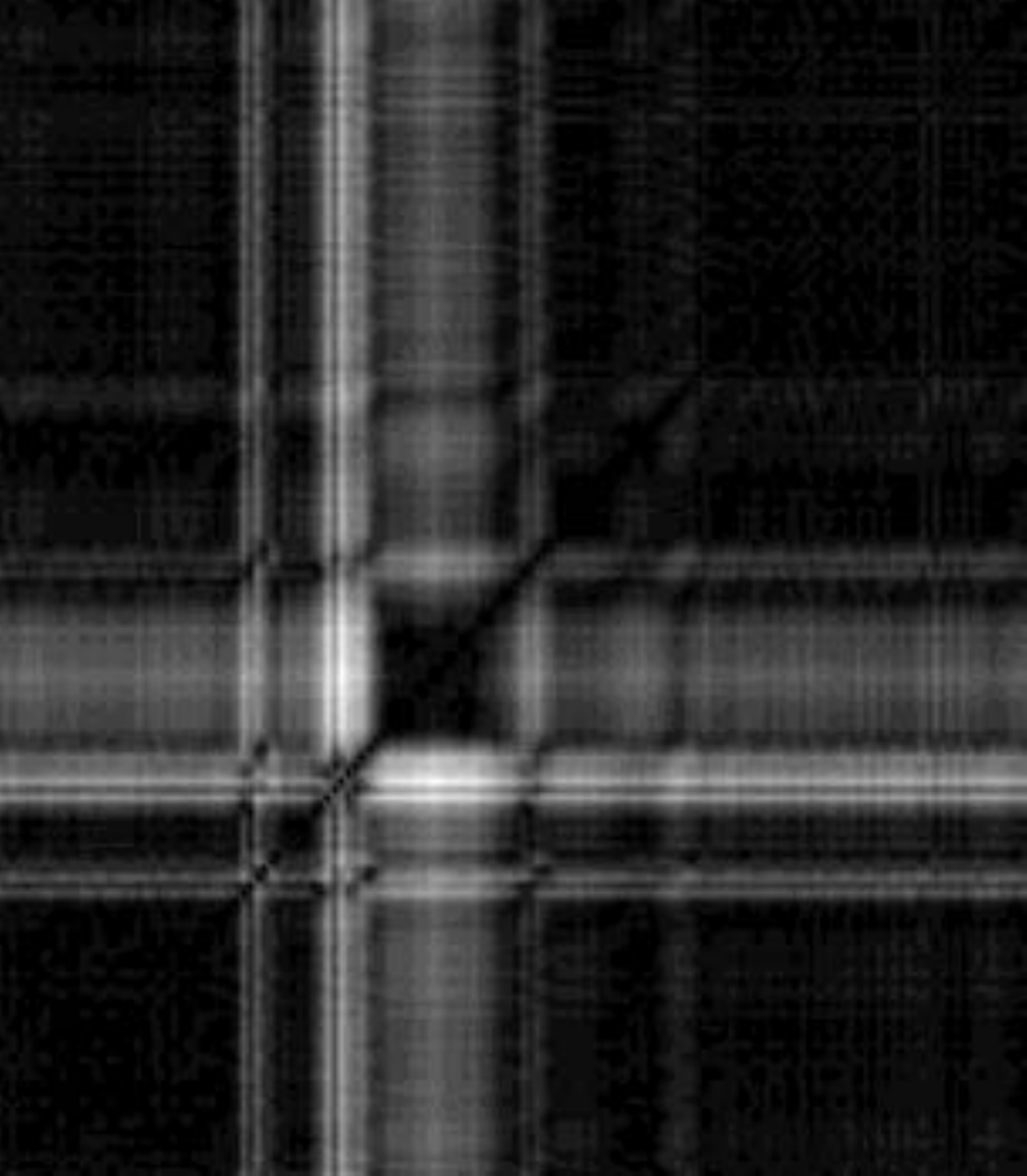} &
          \includegraphics[width=\wRPs]{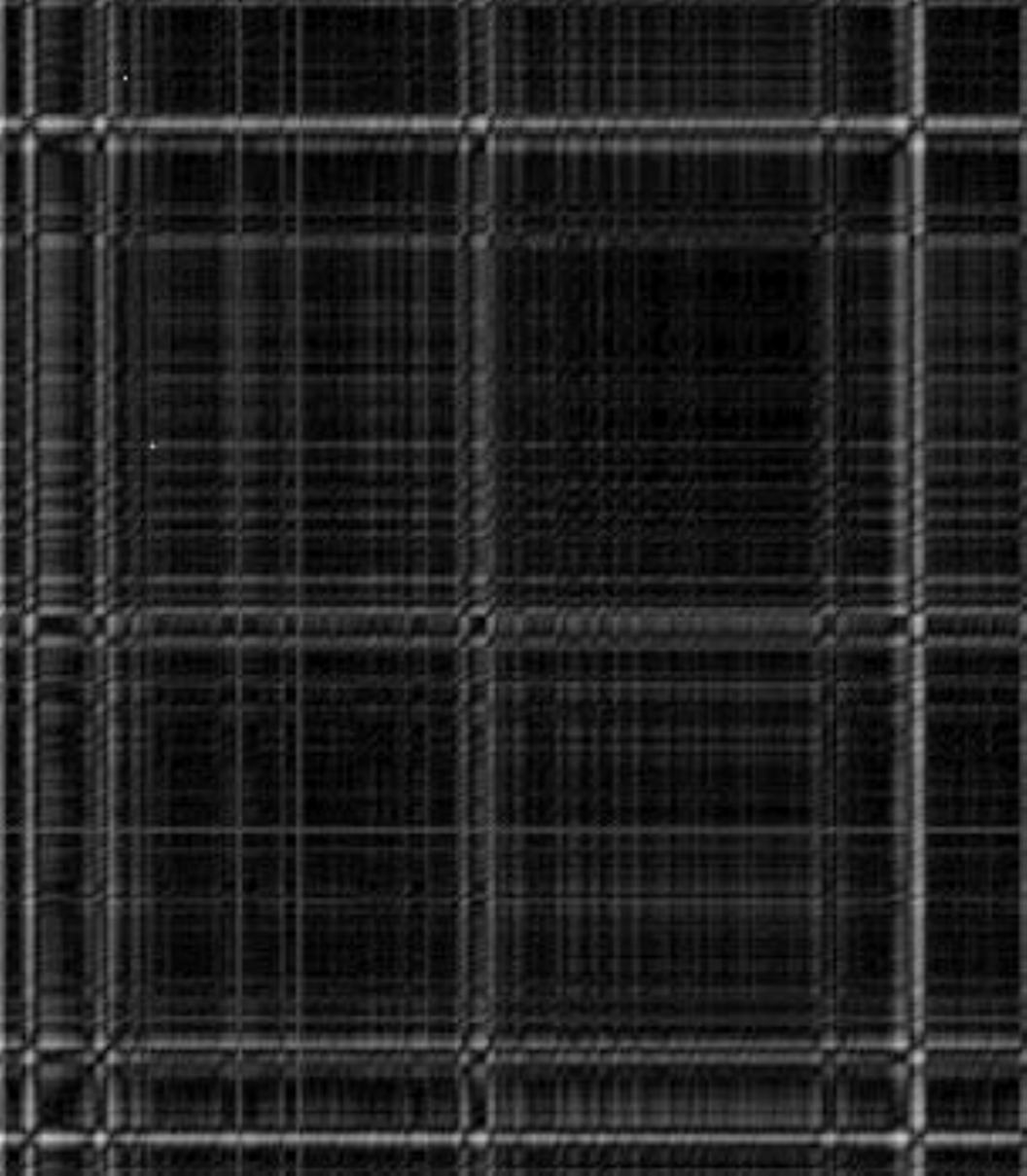} &
          \includegraphics[width=\wRPs]{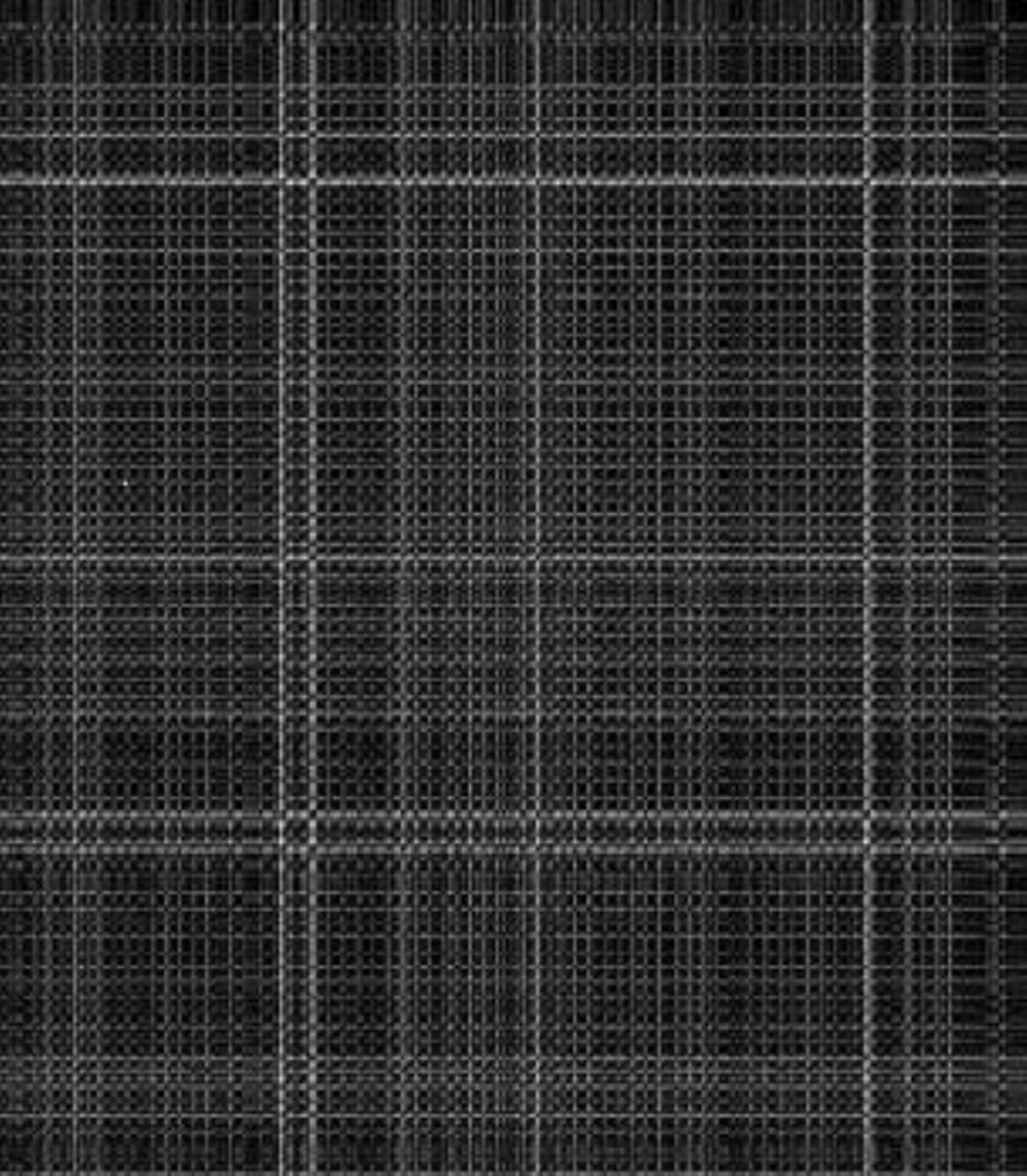} \\

          \includegraphics[width=\wRPs]{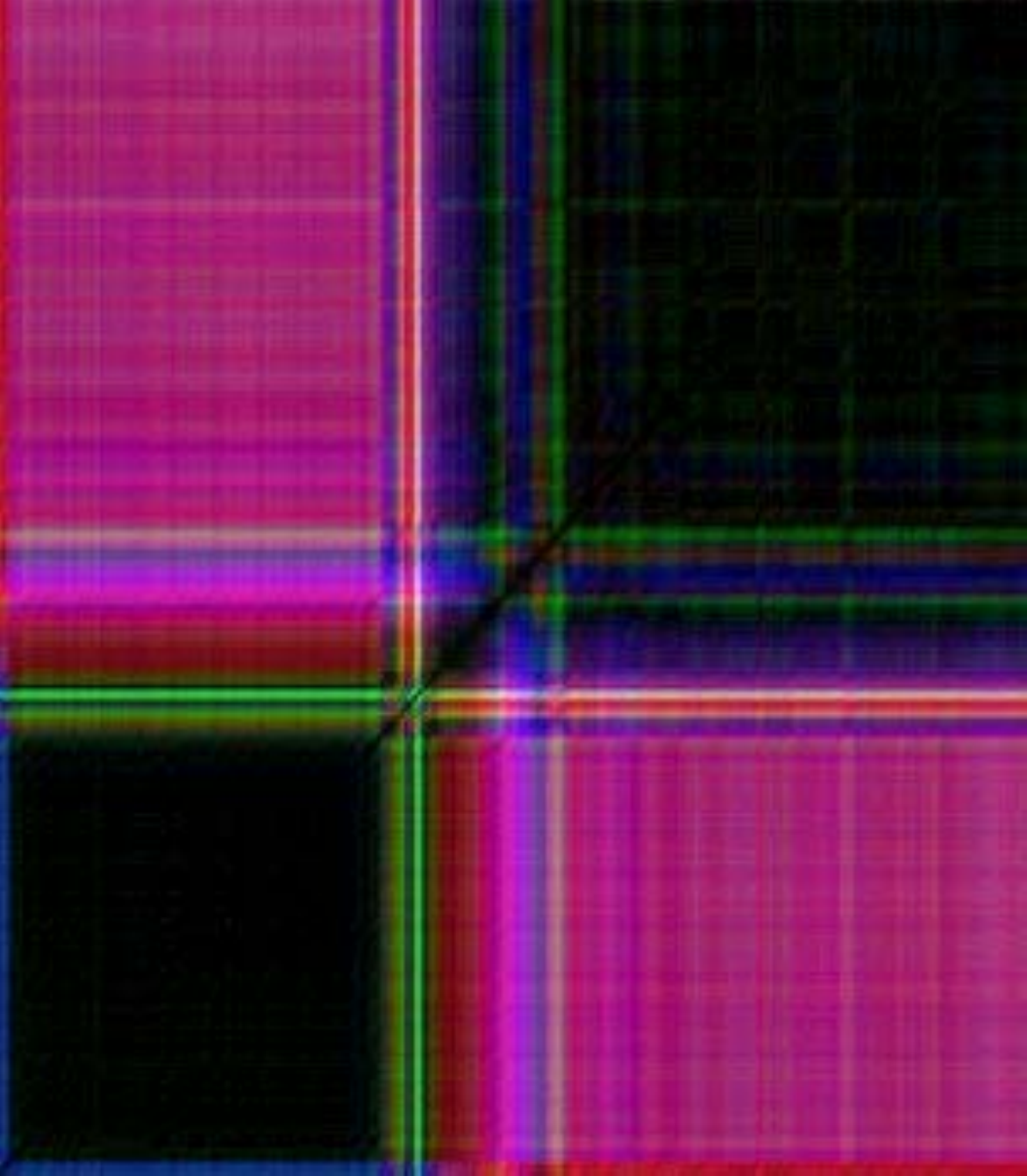} &
          \includegraphics[width=\wRPs]{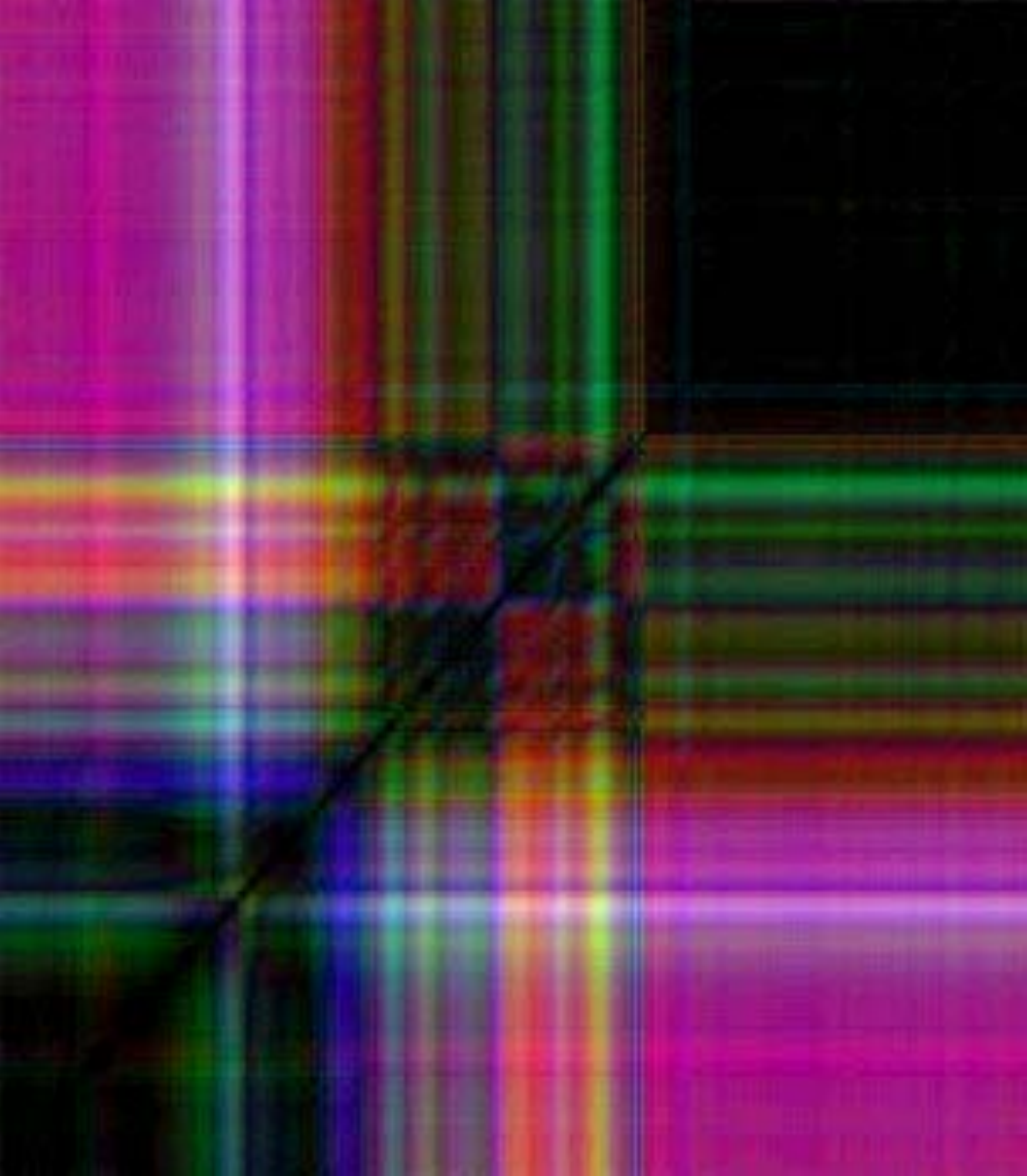} &
          \includegraphics[width=\wRPs]{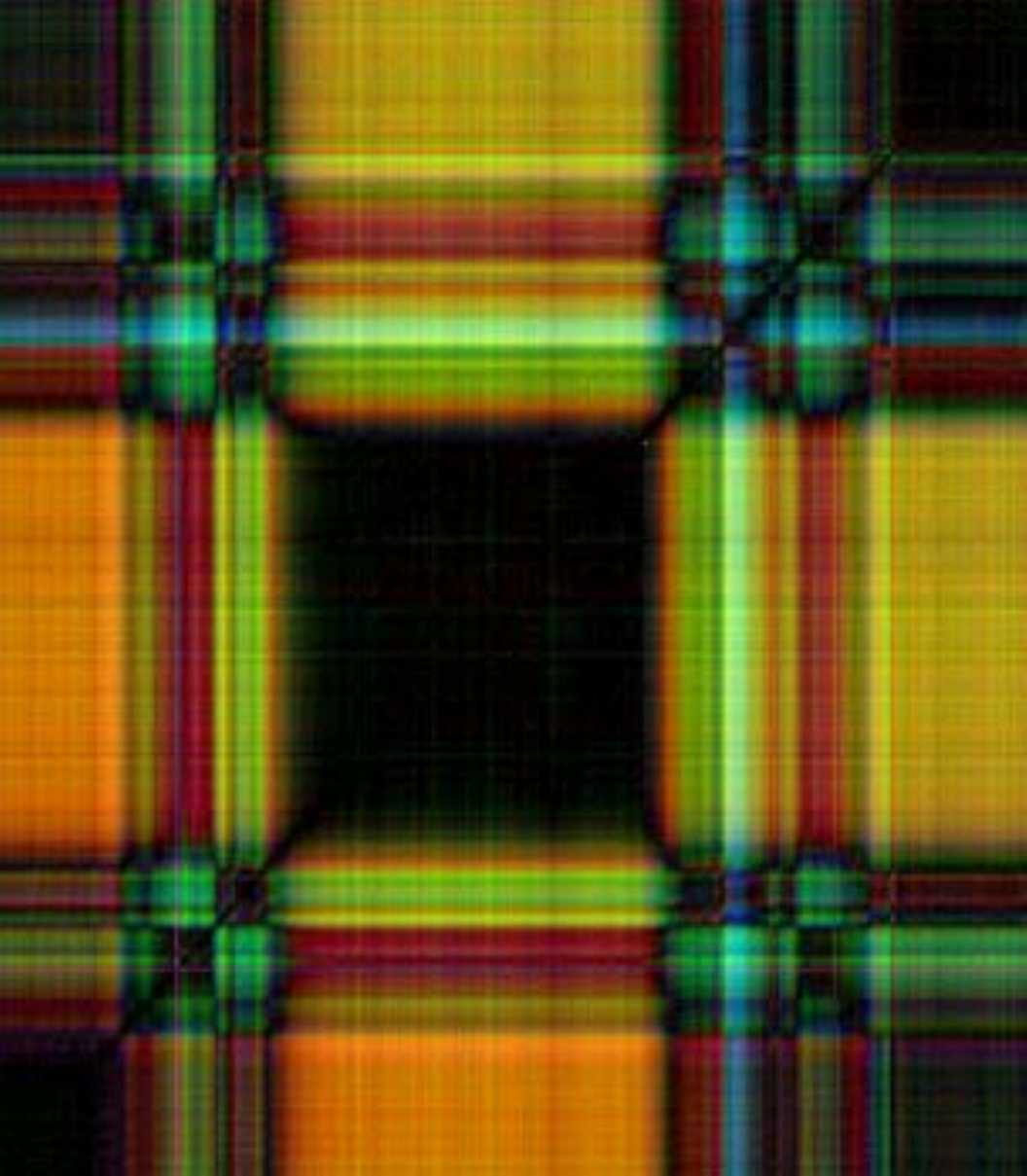} &
          \includegraphics[width=\wRPs]{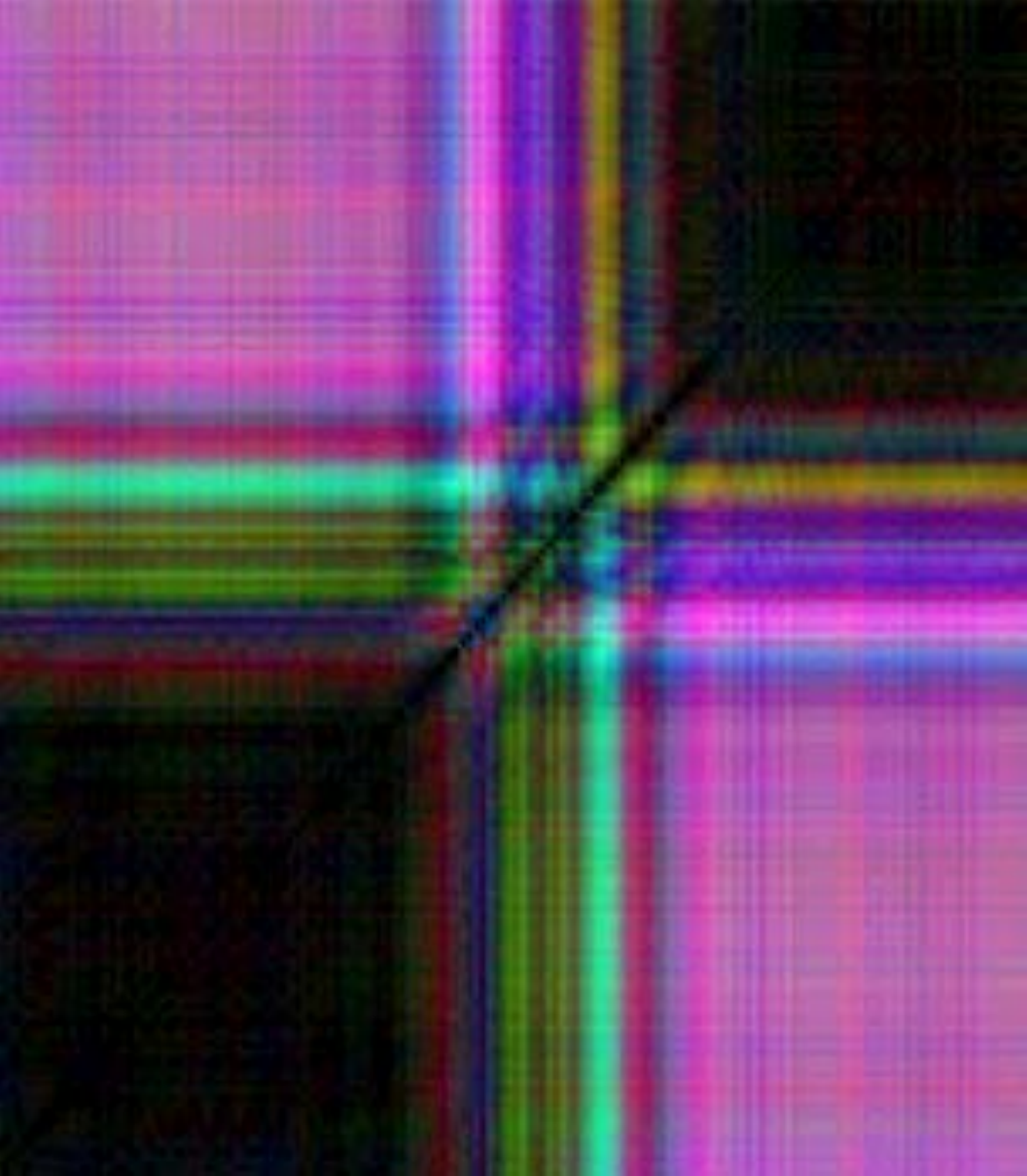} &
          \includegraphics[width=\wRPs]{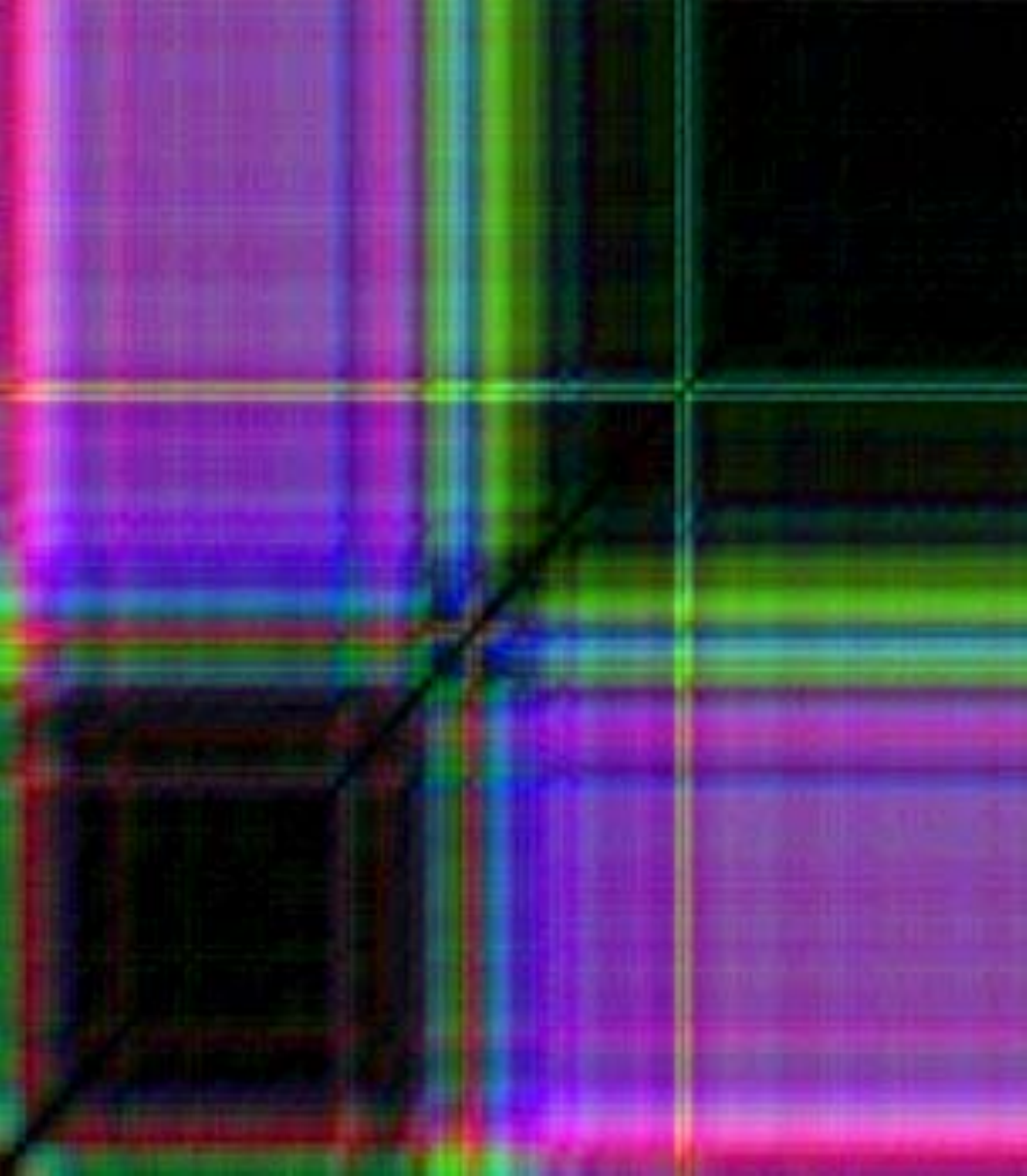} &
          \includegraphics[width=\wRPs]{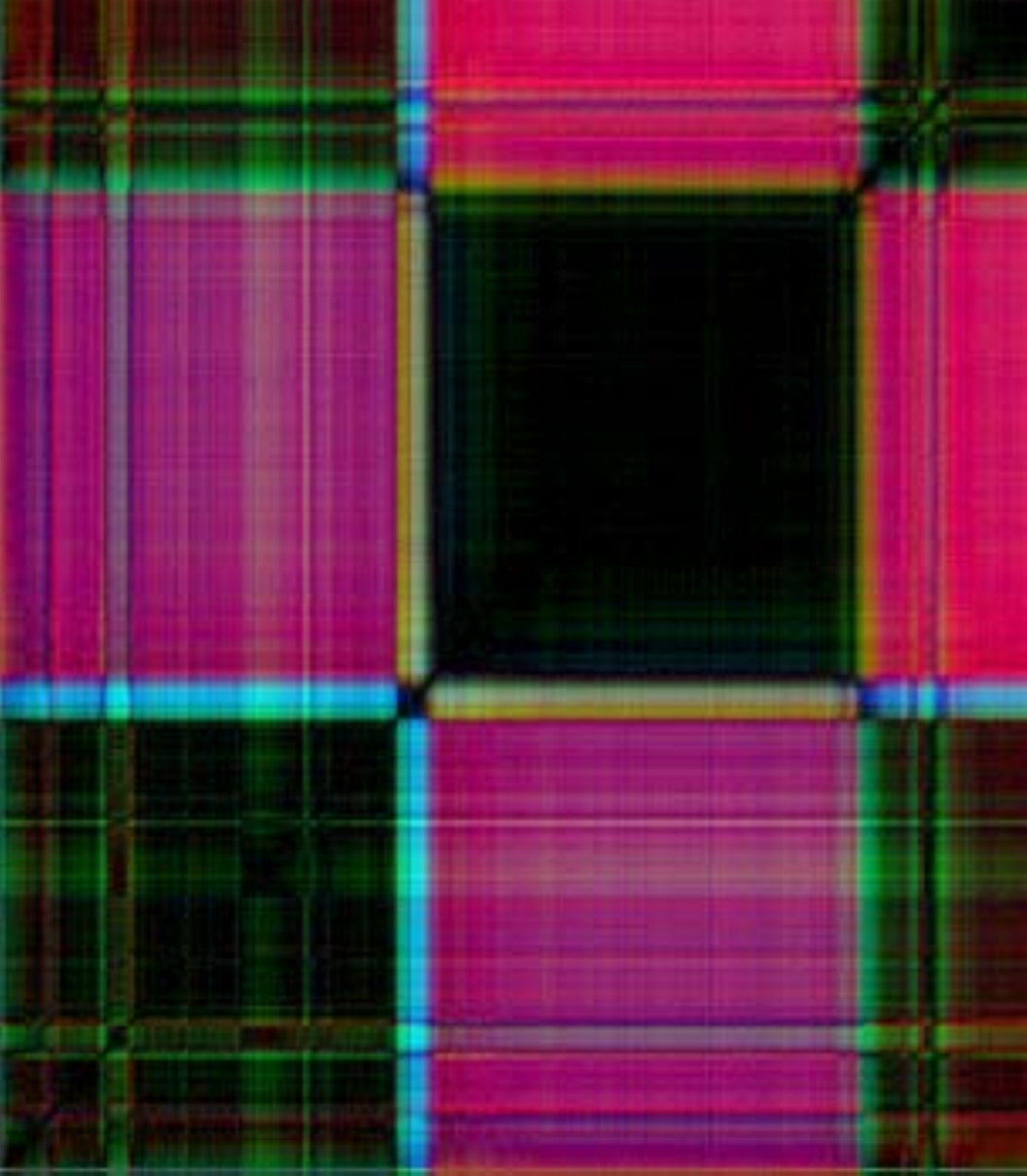} &
          \includegraphics[width=\wRPs]{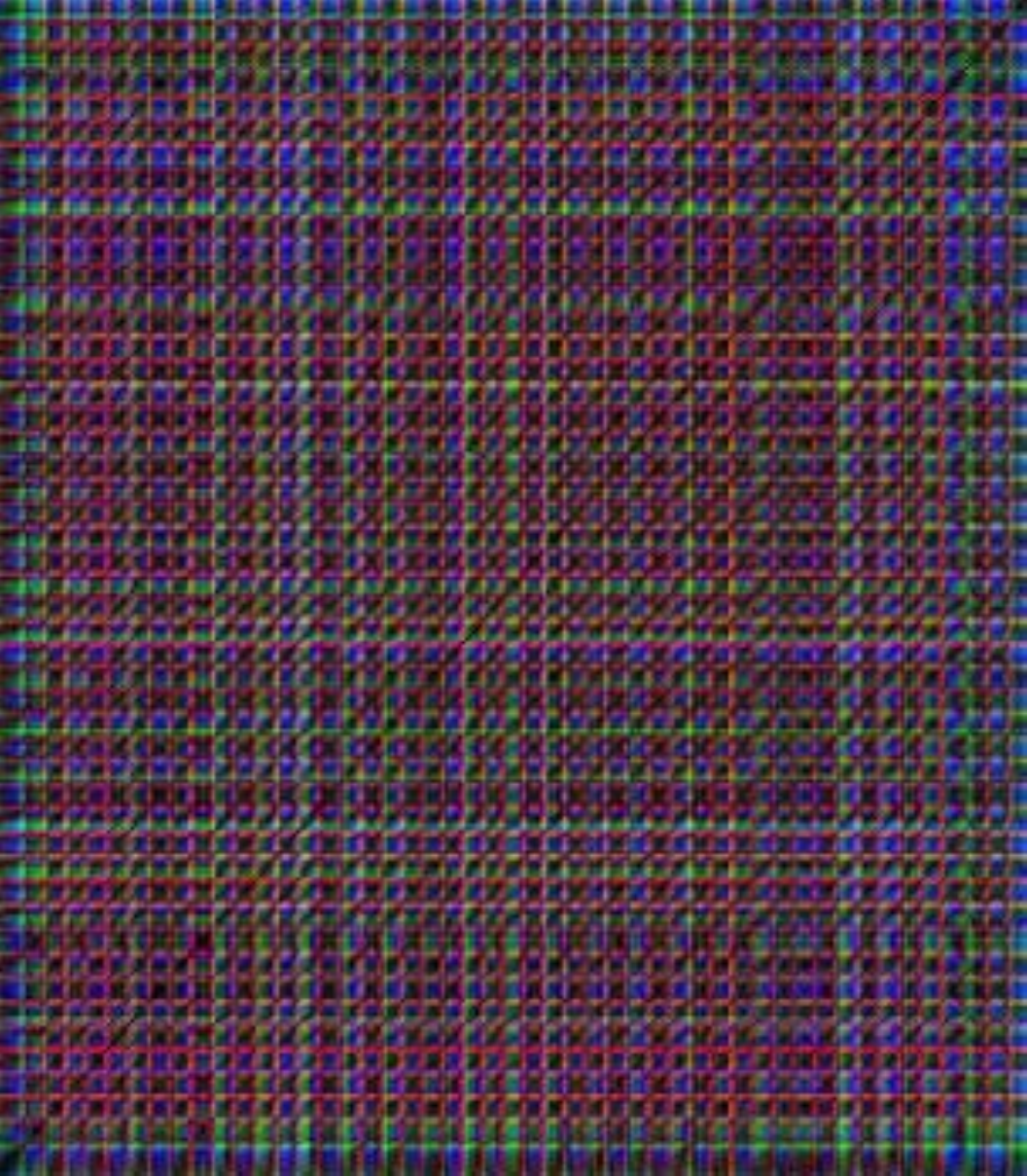} \\

        
        \vspace{-0.3cm}
        \end{tabular}
    \end{center}
    \end{tiny} 
    \caption{Examples of Recurrence Plots (RPs) corresponding to the human activities in the dataset used in our experiments. We can observe the texture patterns of RPs based on accelerometer data and observe the difference among different human activities. Gray-level plots are shown above the RGB plots.}
    \label{fig_examples_rps}
  \end{figure}

Recurrence Plots (RPs) have already been used for many different tasks~\citep{koebbe1992use,Agusti2011,Riva2014,Poh2012,Shohei2007,Dingwell2000,SylosLabini2012,Aboofazeli2008,Holloway2012,Kulkarni2012,Faria2016,Roma2013,Souza2014}, including gait analysis, movement analysis, sound analysis, and also action/activity recognition.
The works that use RPs for HAR~\citep{Kulkarni2012,Holloway2012,Agusti2011} create RPs from videos, not from inertial sensor data, like we are proposing in this paper.
Other recent works~\citep{Poh2012,Kulkarni2012,Souza2014,Faria2016} employ Recurrence Quantification Analysis (RQA) or visual descriptors over RPs and machine learning techniques, but do not consider the use of mobile sensors as the source of information or HAR as the final task.

Our proposal is to employ visual image descriptors for extracting information from RPs and then using such information with machine learning algorithms.
We propose the generation of RPs that take into account the multiple axes of mobile sensor data and then the use of image descriptors that can encode such information effectively.
As RPs are texture images, with our approach, we transform the problem of sensor data classification into a problem of texture classification.

To evaluate the proposed approach, we performed experiments in a dataset of accelerometer data aiming at distinguishing 12 human activities. 
Such activities include brushing teeth, combing hair, drinking glass, pouring water, walking, among others.
As baselines, we considered multiple time- and frequency-domain features.
Results point that the use of RGB Recurrence Plots, in which each color channel represents an axis of the accelerometer data, obtains the highest accuracy levels.
Using Bags of Visual Words (BoVW) with RGB SIFT as the low-level descriptor and max pooling with spatial pyramids is more suitable to encode the small texture differences between RPs of similar activities.

The remainder of this paper is organized as follows.
Section~\ref{related_work} presents related work and 
Section~\ref{proposed_approach} presents the proposed approach.
Section~\ref{experiments} shows experiments and results and 
Section~\ref{conclusions} presents conclusions and future work.

\section{Related work}
\label{related_work}

In this section, we will briefly explain how sensor-based human activity recognition (SB-HAR) is implemented nowadays.
We then focus on existing works that employ Recurrence Plots (RPs) for the analysis and understanding of time-series data, focusing on works similar to our proposal.

SB-HAR~\citep{LabradorHARBook2013,ShoaibSurvey2015} is specially devoted to the understanding of human activities considering only the use of inertial sensor data, like accelerometers and gyroscopes. 
These sensors are present in most smart phones and wearable devices nowadays, making SB-HAR very important for many possible applications, given that people use such devices the whole day and, some of them, even when sleeping. 
Traditionally, SB-HAR is based on the following pipeline: sensor data acquisition, preprocessing, feature extraction, and training of a machine learning classifier from labeled data.  
The learned model is then used to predict the category of new input data~\citep{ShoaibSurvey2015}. 

In this process, the feature extraction step is very important, since the selection of features to be extracted can directly impact the accuracy of the classifier. 
The most common features are time- and frequency-domain features. 
As time-domain features, one can use statistical measures, like mean, standard deviation, root mean square, covariance, histograms, and quantiles. 
Frequency-domain features are based on the Fourier transform, and usually are more costly to compute~\citep{ShoaibSurvey2015}. 
Features computed from the bands of the Fourier transform are commonly used.

After extracting features, the most common classifiers used are Support Vector Machine (SVM), Naive Bayes, K-nearest neighbors (KNN), and Decision Trees~\citep{ShoaibSurvey2015}.

As inertial sensor data can be seen as time-series data, one interesting possibility for SB-HAR is to make use of techniques usually applied to the analysis of time-series data. 
In this field, 
Recurrence Plots are popularly used.

Recurrence Plots (RPs) were proposed in 1987~\citep{EckmannRP1987} and they provide a graphical tool for visualizing the periodic nature of a trajectory through a phase space.
In other words, RPs are a visual representation of time-series data.
They can be defined as~\citep{EckmannRP1987}: 
$$
R\left(i,j\right)=\theta\left(\epsilon-\left\|x_{i}-x_{j}\right\|_{2}\right) \eqno{(1)}
$$

\noindent where $x_{t}$ is the referred time series value in a given time $t$ and $\theta(.)$ is the Heaviside step function. 
In summary, when the trajectory data gets close enough (i.e., within $\epsilon$) to where it has been before, we have a recurrence.
RPs usually look like textures, as we can see in the examples shown in Figure~\ref{fig_examples_rps}.

RPs have been used in many different tasks~\citep{koebbe1992use,Agusti2011,Riva2014,Poh2012,Shohei2007,Dingwell2000,SylosLabini2012,Aboofazeli2008,Holloway2012,Kulkarni2012,Faria2016,Roma2013,Souza2014,Faria2016} like gait analysis, movement analysis, sound analysis, and also action/activity recognition.

One of the most popular ways to extract information from RPs is by using Recurrence Quantification Analysis (RQA)~\citep{Riva2014,Poh2012,SylosLabini2012,Aboofazeli2008,Holloway2012,Roma2013}.
RQA quantifies the properties of dynamical systems like the number and duration of recurrences based on the phase space trajectory.
RQA can quantify the small-scale structures of RPs using different types of measures.
Popular RQA measures are: recurrence rate, determinism, laminarity, trapping time, divergence, etc.

Our approach is based on using image descriptors over RPs instead of applying RQA.
Comparing RQA and image descriptors, we can say that RQA measures are usually a single value, which in many cases cannot capture the spatial arrangement of certain texture properties of RPs.
Image descriptors, on the other hand, are multidimensional vectors (usually lying on $\mathbb{R}^{d}$ with $d > 100$) and can be very precise in capturing such properties~\citep{PenattiWSAPR2014}.

Poh et al.~\citep{Poh2012} used RQA to extract features from RPs with SVM as classifier aiming at classifying accelerometer data of epileptic seizures.

Kulkarni and Turaga~\citep{Kulkarni2012} used non-thresholded recurrence matrices as distance matrices and extracted an image descriptor (Local Binary Patterns - LBP) to encode their textural properties. 
Such features were used with a nearest neighbor classifier to classify videos of human activities captured by compressive cameras.

The work of Souza et al.~\citep{Souza2014} proposes the use of image texture descriptors for extracting information from RPs aiming at classifying time-series data.
They use SVM as classifier and consider the tasks in the UCR Time Series Archive~\citep{UCRArchive}. 
This work is similar to ours, however, they consider only gray-level plots, while we use RGB plots in which each dimension refers to an axis of the accelerometer data.

Faria et al.~\citep{Faria2016} proposed the use of image descriptors over RPs for plant recognition.
They obtained time-series data from images taken of plants over time. 
Such images are segmented into regions and the regions are considered in terms of how they contribute to each R, G, and B channels separately.
The average pixel intensity is then computed per channel and the variation of this value over time is used as source for computing the time series.
As they have one time series per color channel, they compute a recurrence plot for each channel and then combine the three plots into a single RGB plot.
After that, they extract image descriptors from the RGB plot and use them in a classification protocol.
Although this work is similar to ours, we are using data from mobile inertial sensors (e.g., accelerometers) as source for the time-series data.
We are also considering the task of activity recognition and we are using image descriptors more suitable to the description of RGB RPs, as we can encode the spatial differences between RPs.
We show that the descriptors used in our work obtain higher accuracy than some of the descriptors used in~\citep{Faria2016}.

\section{Proposed approach}
\label{proposed_approach}

\begin{figure}[t!]
    \begin{center}
      \includegraphics[width=\columnwidth]{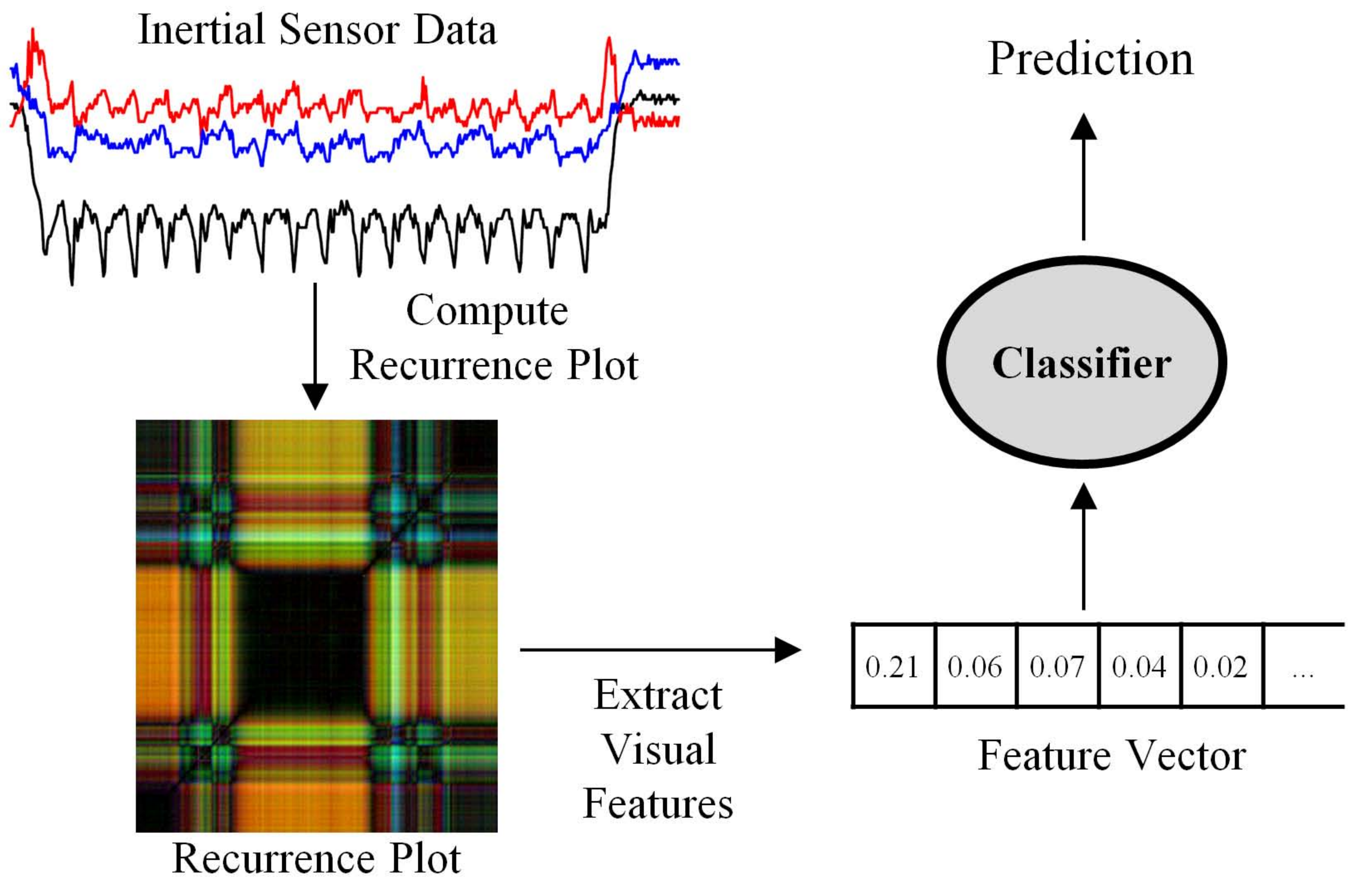} \vspace{-0.3cm}
    \end{center}
    \caption{Scheme of the proposed approach: Recurrence Plots are created from inertial sensor data and then used with computer vision techniques for human activity recognition.}
    \label{fig_method}
  \end{figure}

The proposed approach, illustrated in Figure~\ref{fig_method}, is based on the use of Recurrence Plots (RP) computed from inertial sensor data and then using computer vision techniques for human activity recognition.
The main intuition is that the visual analysis of recurrence plots can provide discriminating information for HAR, being able to distinguish human activities measured by sensor data.
The different visual patterns in recurrence plots for different human activities can be noted in the examples shown in Figure~\ref{fig_examples_rps}.

Recurrence Plots are mainly based on a single time series, however, sensor data is usually composed of several axis.
For instance, accelerometer and gyroscope data, which are common sensors available in mobile devices, have three axes: x, y, and z.
RPs need to be computed separately for each axis and then combined to have a single RP for each input data. 
Another option is to combine/summarize the sensor data axes before computing the RP, thus obtaining a single RP for each input data.
In our approach, we explore different possibilities in this sense.
A first option (i) is to compute the vector norm of the 3D sensor data (x,y,z) and then compute the RP from the vector norm, generating a gray-level RP.
Another option (ii) is to compute a separate RP for each axis of the data and then concatenate the RPs side by side, still generating a gray-level plot, except that it is wider than the first option.
Other option (iii) is to compute a separate RP for each axis and then use each RP as a channel in the RGB color space.
That is, the RP for the x-axis will be the R channel of the final RP, y-axis will be the G channel and z-axis, the B channel.
This results in an RGB plot.
These three possibilities will be named respectively as: (i) RP-gray, (ii) RP-gray-concat, (iii) RP-RGB.

After computing RPs, computer vision approaches are employed in order to ``understand'' the visual content from RPs.
As we can see in some examples of RPs in Figure~\ref{fig_examples_rps}, they look like textures.
Therefore, with the proposed approach of using RPs for HAR, we transform the activity recognition problem into a texture recognition problem.

For extracting features from RPs, one can employ any of the existing feature descriptors used for image understanding, like descriptors based on gradient information (e.g., Histogram of Oriented Gradients - HOG~\citep{DalalHOGCVPR2005} or Scale-invariant Features Transform - SIFT~\citep{LoweSIFT2004}), descriptors based on visual codebooks~\citep{BoureauMidLevelCVPR2010,GemertVisuaWordAmbiguityPAMI2010} and also descriptors based on Convolutional Neural Networks (CNN)~\citep{OverFeatIntegrated2014}.
As RPs have texture patterns, one should prefer descriptors devoted to encode this property.

The extracted features are then used to train a machine learning classifier.
The learned model during training is then used for predicting the category of unknown input data, when the system is being used.

\section{Experiments and results}
\label{experiments}

To evaluate the proposed approach, we performed experiments in a dataset of accelerometer data using a classification protocol, reporting results in terms of the classification accuracy for human activity recognition. 

The dataset used, called WHARF~\citep{BrunoWHARF2015}, contains data recorded by a tri-axial accelerometer sensor used in the right wrist of subjects.
The sensor range is of [-1.5G; +1.5G] with sensitivity of 6 bits per axis and sampling frequency of 32Hz.
The complete dataset has 892 samples divided into 14 classes of human activities, as shown in Table~\ref{tab_wharf}. 
Given the small number of samples for the `eating' classes, they were discarded in our experiments. 
Therefore, the final dataset used in the experiments has 884 samples of 12 classes.

\begin{table}[h]
	\centering
        \caption{Details about the dataset used in the experiments: classes and number of samples per class. }
		\begin{tabular}{rlr} \hline
            \# &\textbf{Class} & \textbf{Size} \\ \hline
            1 & brush teeth    &  12 \\
            2 & climb stairs   & 112 \\
            3 & comb hair      &  31 \\
            4 & descend stairs &  51 \\
            5 & drink glass    & 115 \\
            6 & eat meat       &   5 \\
            7 & eat soup       &   3 \\
            8 & get up bed     & 101 \\
            9 & lie down bed   &  28 \\
            10 & pour water     & 100 \\
            11 & sit down chair & 109 \\
            12 & stand up chair & 112 \\
            13 & use telephone  &  13 \\
            14 & walk           & 100 \\ \hline
		\end{tabular}
        \label{tab_wharf}
\end{table}

As baselines, we used traditional time- and frequency-domain features, extracted directly from the accelerometer data.
As time-domain features, we selected the following statistical measures: mean, standard deviation, quantile (21 probabilities), histogram (16 bins per axis), root mean square, and covariance (Pearson). 
Many other measures were also tested (e.g., max, min, number of peaks, correlation, entropy, other covariance measures, mean absolute deviation, kurtosis, skewness), but as they obtained lower accuracy rates, we decided to not report their results.
We also used frequency-domain features based on the bands of the Fast Fourier transform (fftbands) and Discrete Fourier transform (dftbands).
The fftbands were computed by using \textit{fft} function of \textit{stats} package on R~\citep{R2008} and splitting the magnitude vector in 10 bands with same size (except by the last band, that was truncated). 
The dftbands were computed by regular exponential calculation and splitting in 10 bands in the same way of fftbands.
We are not considering the fusion of multiple features, as the objective is to evaluate the discriminating power of each feature independently.

In the proposed approach, we evaluated three types of RPs, as explained in Section~\ref{proposed_approach}: 
\begin{itemize}
	\item RP-gray: gray-level from vector norm, 
    \item RP-gray-concat: gray-level concatenating the RP of each axis side by side, and
    \item RP-RGB: RP of each axis as one RGB channel.
\end{itemize}
The RPs were created with the \textit{recurr} function (parameters `m' and `d' with value 2) from \textit{tseriesChaos} package in R~\citep{R2008}.

For extracting visual information from the RPs, we evaluated the following visual descriptors: LAS~\citep{TaoLASJVCI2000}, SASI~\citep{SASIPR2003}, BIC~\citep{StehlingBIC2002}, HOG~\citep{DalalHOGCVPR2005}, features from OverFeat (deep convolutional neural network, last layer before the soft-max layer of the fast model)~\citep{OverFeatIntegrated2014}, and several configurations of bags of visual words (BoVW)~\citep{BoureauMidLevelCVPR2010,GemertVisuaWordAmbiguityPAMI2010,SandeEvaluatingTPAMI2010,LazebnikSPM2006}.
For BoVW, we considered dense sampling (6 pixels) in all cases~\citep{SandeEvaluatingTPAMI2010}.
As low-level descriptors, we used SIFT for the gray-level plots and RGB Histogram, OpponentSIFT and RGB SIFT for the RGB plots.
Visual codebooks were created by random selection, having from 100 to 10000 visual words.
To create the BoVW vector of each RP, we used hard and soft assignment (codeword uncertainty with $\sigma$=150)~\citep{GemertVisuaWordAmbiguityPAMI2010} and average pooling, max pooling~\citep{BoureauMidLevelCVPR2010} or max pooling with spatial pyramids (maxSPM)~\citep{LazebnikSPM2006}.

The classification protocol is the following: random selection of $10$ samples per class for training and using the rest for testing.
This guarantees a balanced training set in terms of samples per class.
Linear Support Vector Machine (SVM) was used as classifier.
Given the random parts of the protocol, every experiment was run 10 times and results are reported by the normalized average accuracy and confidence intervals for 95\% of confidence ($\alpha$=0.05) for the 10 runs. 
We also show a confusion matrix for the best performing methods.

\begin{figure}[t!]
    \begin{center}
      \includegraphics[width=0.45\textwidth]{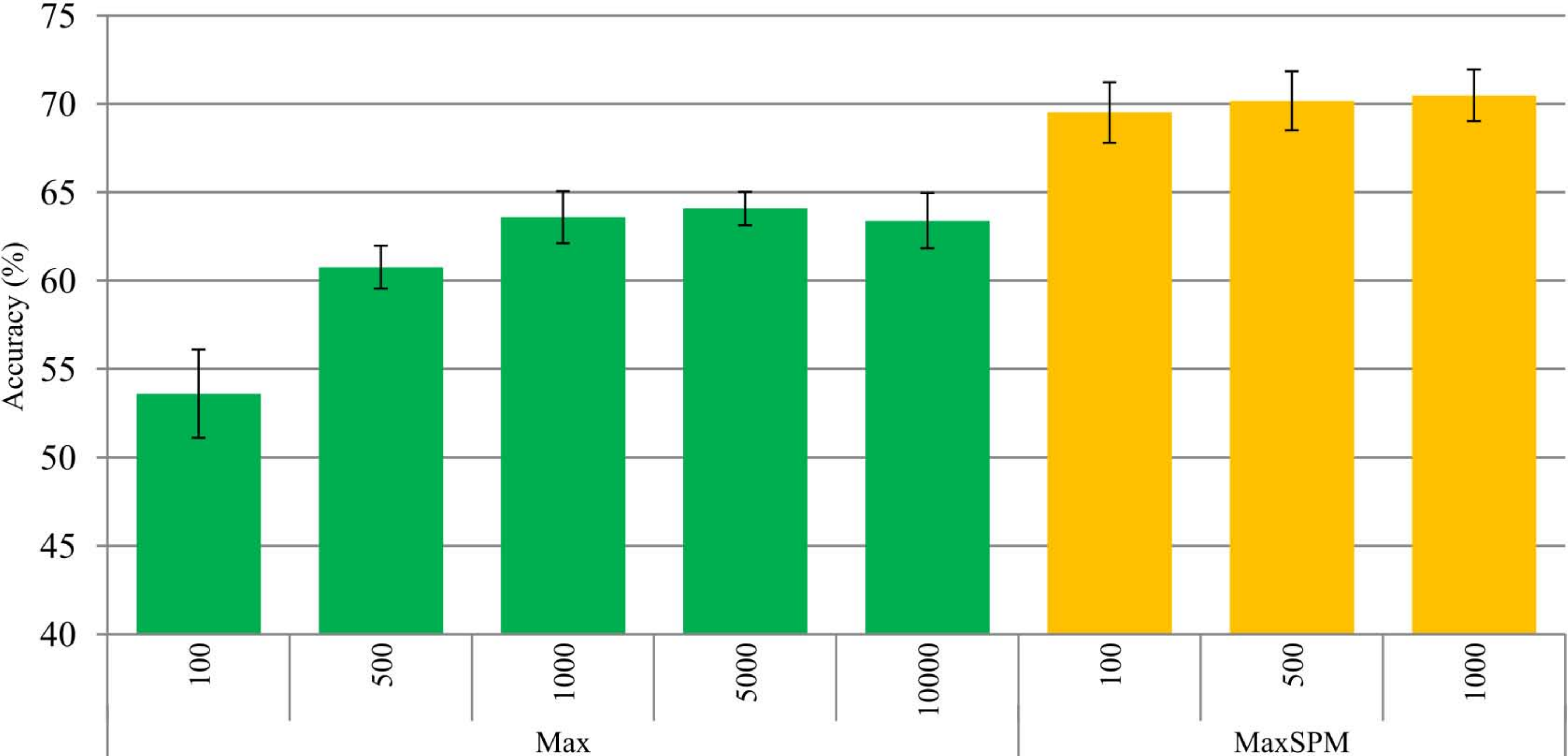} \vspace{-0.3cm}
    \end{center}
    \caption{Evaluation of BoVW configurations for the gray-level recurrence plots varying the codebook size and the pooling method. We can note that codebooks up to 1000 visual words obtain improvements in accuracy. We can also see that spatial pyramids (maxSPM) are very important for higher accuracy levels. The results consider dense SIFT as the low-level descriptor.}
    \label{fig_results_bovw_gray}
  \end{figure}

Before comparing the visual descriptors with the other baseline features, we performed an evaluation of the parameters involved in the BoVW features.
We initially evaluated BoVW descriptors in the recurrence plots computed from the vector norm of the tri-axial accelerometer data, i.e., BoVW features over the gray-level plots (RP-gray).
Figure~\ref{fig_results_bovw_gray} shows the impact of different codebook sizes and the impact of spatial pyramids (SPM) in the results.
We can see that by increasing the codebook size from 100 to 500 and to 1000 visual words, we obtained considerable improvements. 
Larger codebooks (5000 and 10000) did not represent considerable gains in accuracy.
In fact, the use of large codebooks can harm efficiency, because feature vector sizes also increase.
For instance, when using the codebook of 1000 visual words, we have a feature vector of 21000 dimensions considering spatial pyramids of 2 levels.
For a codebook of 10000 words, it would be 10 times larger, i.e., 210 thousand dimensions.
We are not showing the results when using hard assignment and average pooling, as they usually achieved lower accuracy rates.

Figure~\ref{fig_results_bovw_gray} also shows improvements of around 10 percentage points when using spatial pyramids, for most codebook sizes tested. 
The reason is that for many activities, textures are similar and the difference relies on the location of specific texture patterns. 
To encode such differences, spatial pooling methods, like spatial pyramids, are important for better results.
Therefore, in the following experiments, we report the BoVW results using the codebook of 1000 visual words and max pooling with spatial pyramids (maxSPM).

\begin{figure}[t!]
    \begin{center}
      \includegraphics[width=0.4\textwidth]{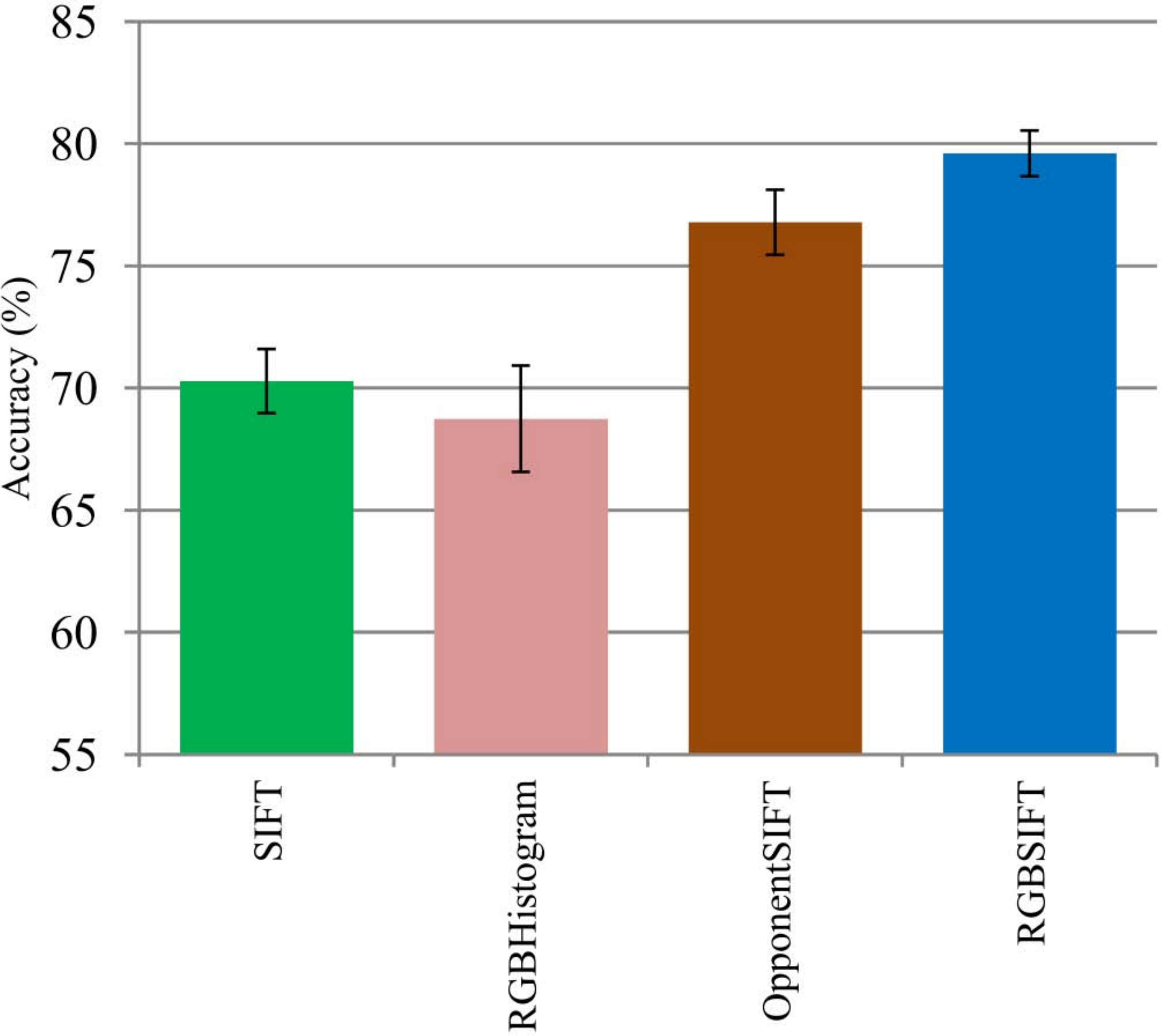} \vspace{-0.3cm}
    \end{center}
    \caption{Analysis on BoVW features for the RGB plots using different low-level color descriptors. We can note that color information improves the results, with RGB SIFT obtaining the highest accuracies. BoVWs here are based on 1000 visual words and soft assignment with $\sigma=150$, except for RGB Histogram, which was better with 100 visual words and soft assignment with $\sigma=0.2$.}
    \label{fig_results_bovw_color}
  \end{figure}

For RGB plots, we also have the option to choose different low-level color features.
Hence, we performed an evaluation of which color descriptor would be more appropriate.
We evaluated RGB Histogram, OpponentSIFT and RGB SIFT. 
As a matter of comparison, we also show the results of the usual gray-level SIFT descriptor even using the RGB plots.
Figure~\ref{fig_results_bovw_color} shows the clear gains in accuracy when using color descriptors in comparison to the gray-level SIFT descriptor.
RGB Histogram, however, has not improved accuracy over SIFT.
As color histograms, like RGB Histogram, usually have low discriminating power because they do not encode texture information, even gray-level SIFT was better.
RGB SIFT and OpponentSIFT obtained the highest accuracy rates, with RGB SIFT being superior.
As RGB SIFT works by applying the SIFT descriptor for each RGB channel independently, it could better encode the texture properties of each accelerometer axis.

\begin{figure*}[t!]
    \begin{center}
      \includegraphics[width=0.9\textwidth]{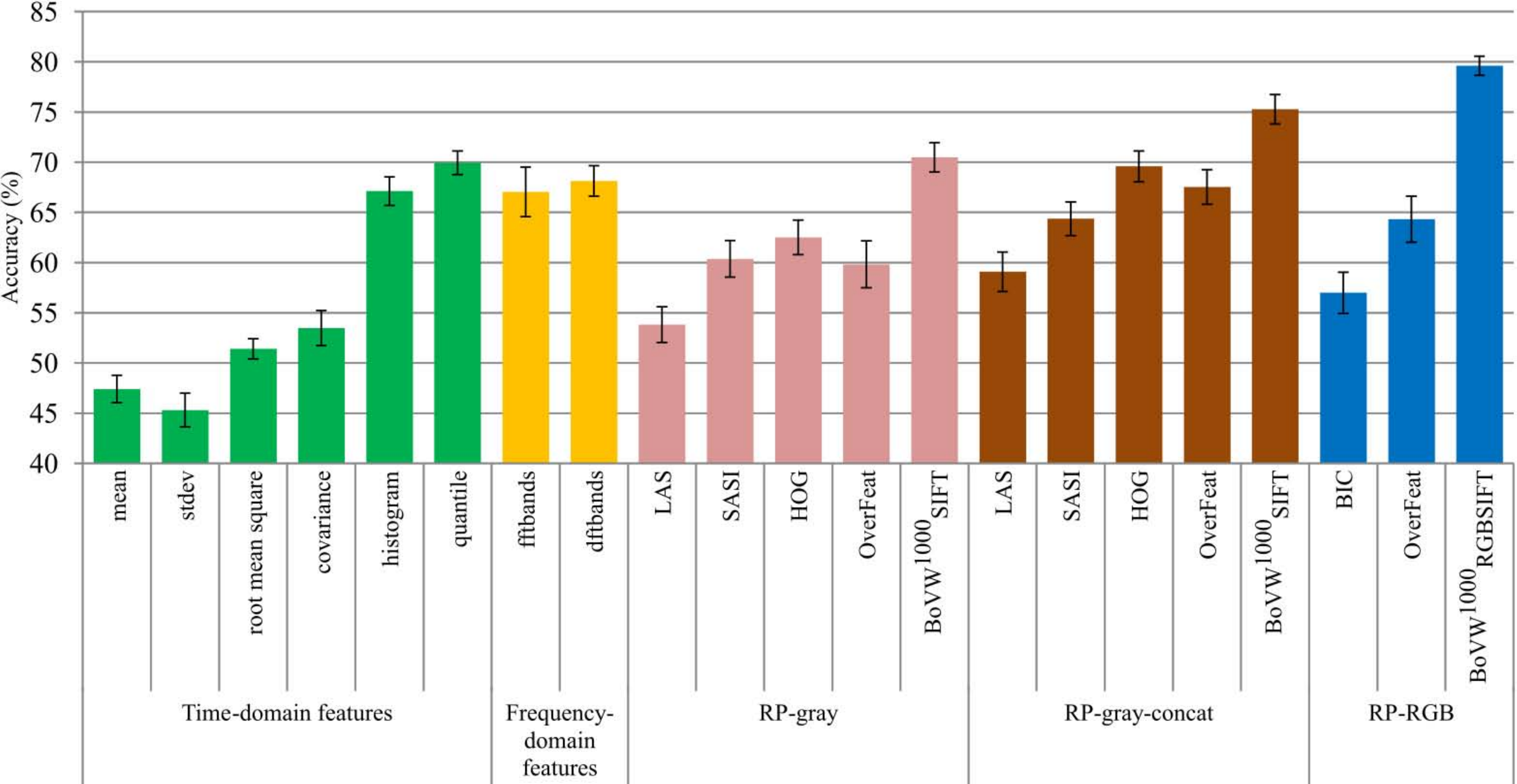} \vspace{-0.3cm}
    \end{center}
    \caption{Classification results: the highest accuracies are obtained by the proposed approach using the RGB recurrence plots.}
    \label{fig_results}
  \end{figure*}

Figure~\ref{fig_results} shows the results comparing the baselines with the visual features extracted from recurrence plots. 
We can observe that the use of visual descriptors over RPs outperforms the features directly extracted from accelerometer data.
The best time- and frequency-domain features achieve around 70\% of classification accuracy, while the best visual features for the RGB plots achieved almost 80\%.

Comparing the specific types of features, we can note that quantile and histogram are the best time-domain features, with comparable results to the frequency-domain features tested.
In fact, quantile achieves comparable results with the BoVW descriptors for gray-level RPs.
Comparing the results for the gray-level RPs, we can see that BoVW features outperform the others.

Comparing the two types of gray-level plots (RP-gray and RP-gray-concat), the accuracy is higher when accelerometer axes are considered independently (i.e., RP-gray-concat), by not computing the vector norm before the RP.

The features extracted by OverFeat, which is a convolutional neural network (CNN) trained to recognize the 1000 object categories of ImageNet, were not as discriminant as BoVW features.
As objects are very different from the texture patterns of recurrence plots, we observed that such features are not adequate in this case. 
We believe that a CNN trained specifically for discriminating textures would obtain better results.

\begin{figure}[b!]
    \begin{center}
      \includegraphics[width=0.3\textwidth]{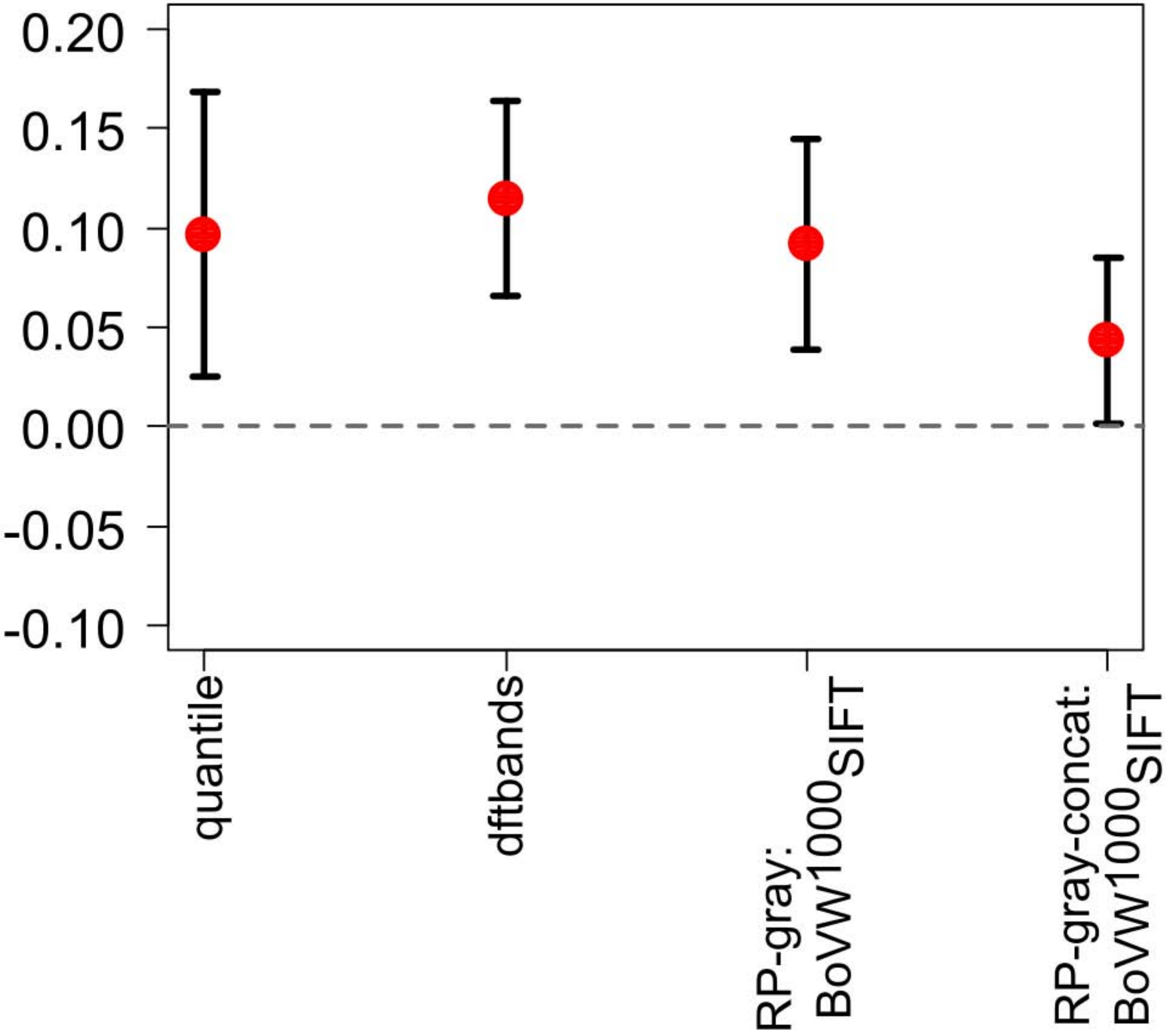} \vspace{-0.3cm}
    \end{center}
    \caption{Paired test comparing RP-RGB $BoVW^{1000}_{RGBSIFT}$ to the best method of each modality. We can see that, as all intervals are above zero, $BoVW^{1000}_{RGBSIFT}$ is statistically better than each other method presented in the horizontal axis.}
    \label{fig_paired_test}
  \end{figure}

To verify statistical significance in the results, we performed a paired test, comparing the best performing BoVW feature ($BoVW^{1000}_{RGBSIFT}$ for the RGB plots) against the best performing method of each kind: quantile, dftbands, RP-gray $BoVW^{1000}_{SIFT}$, and RP-gray-concat $BoVW^{1000}_{SIFT}$. 
To perform this statistical test, we computed the average accuracy per class among the 10 runs for each method. 
Then, we computed the pair-wise per class difference between the two methods being compared and computed the average of the differences with confidence intervals (for confidence level of 95\%).
If the confidence interval crosses zero, there is no statistical difference between methods.
If the difference is above zero, the first method is statistically better than the second. 
Otherwise, the second method is better.
Figure~\ref{fig_paired_test} shows this analysis graphically. 
We can see that, in all cases, $BoVW^{1000}_{RGBSIFT}$ is superior to the other methods.

\begin{figure}[t!]
    \scriptsize
	\begin{center}
        \begin{tabular}{@{}c@{ }c@{}}
            quantile & dftands \\
            \includegraphics[width=0.24\textwidth]{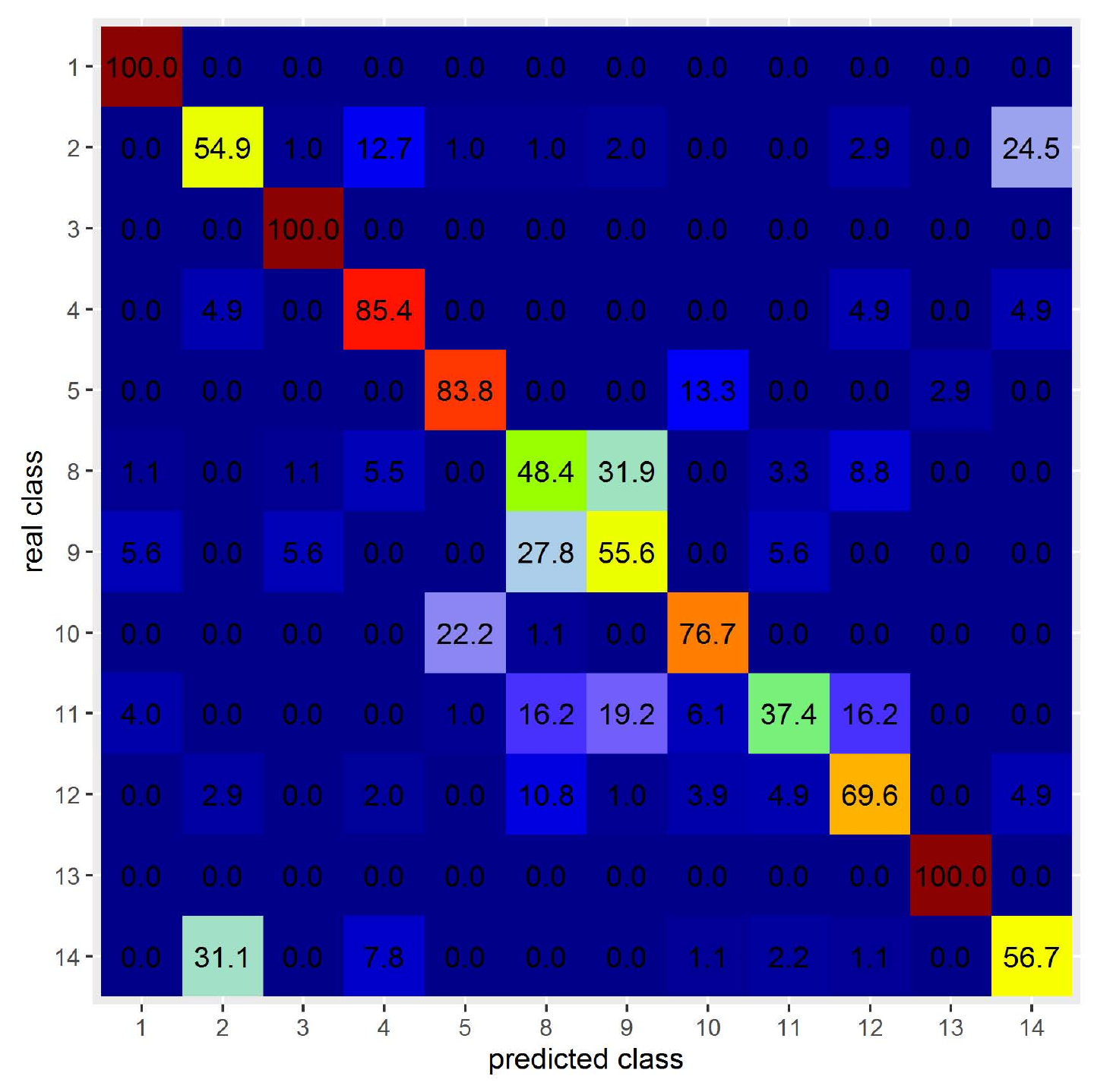} &
            \includegraphics[width=0.24\textwidth]{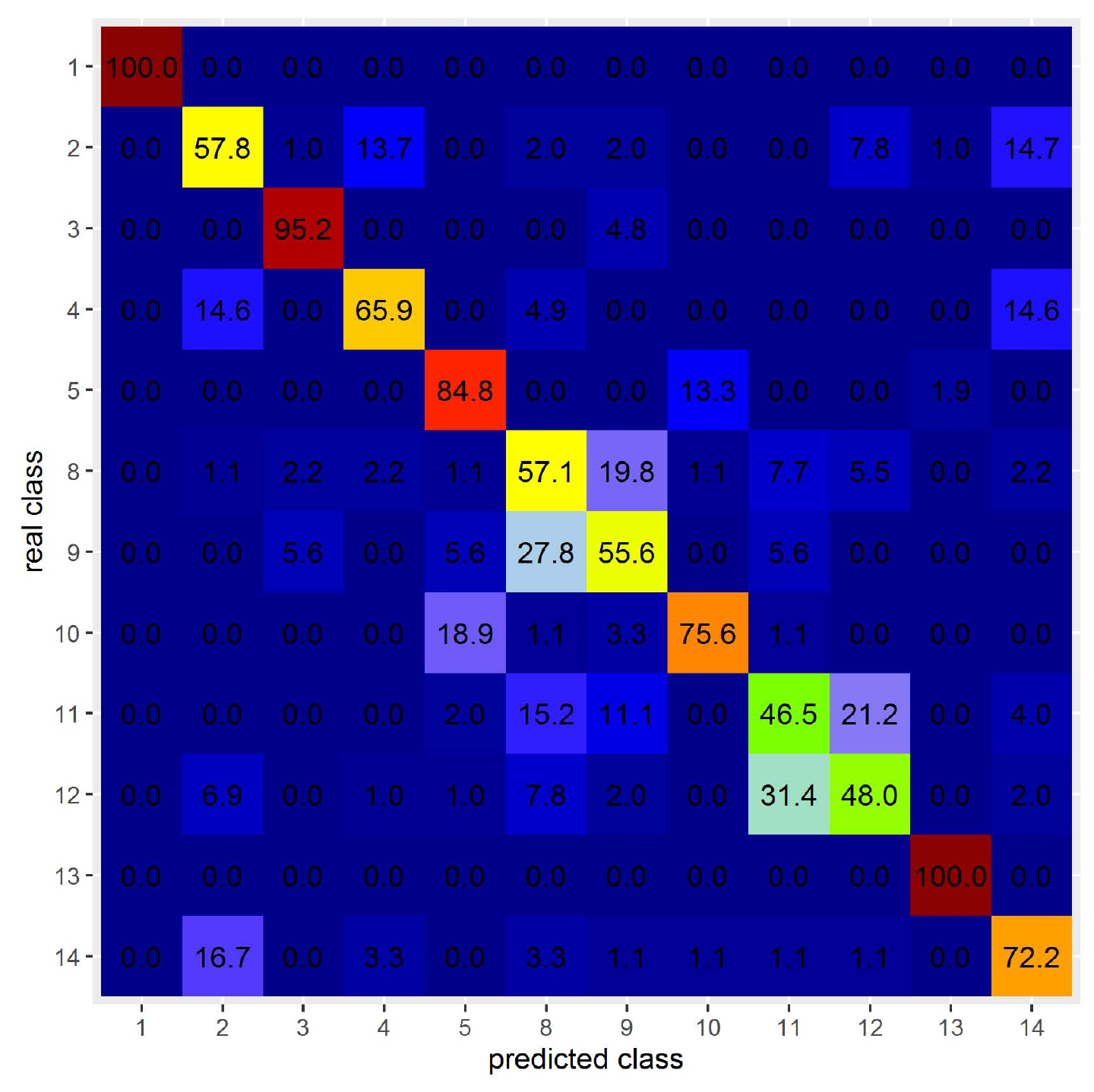} \vspace{0.5cm} \\

            RP-gray $BoVW^{1000}_{SIFT}$ & RP-gray-concat $BoVW^{1000}_{SIFT}$ \\
            \includegraphics[width=0.24\textwidth]{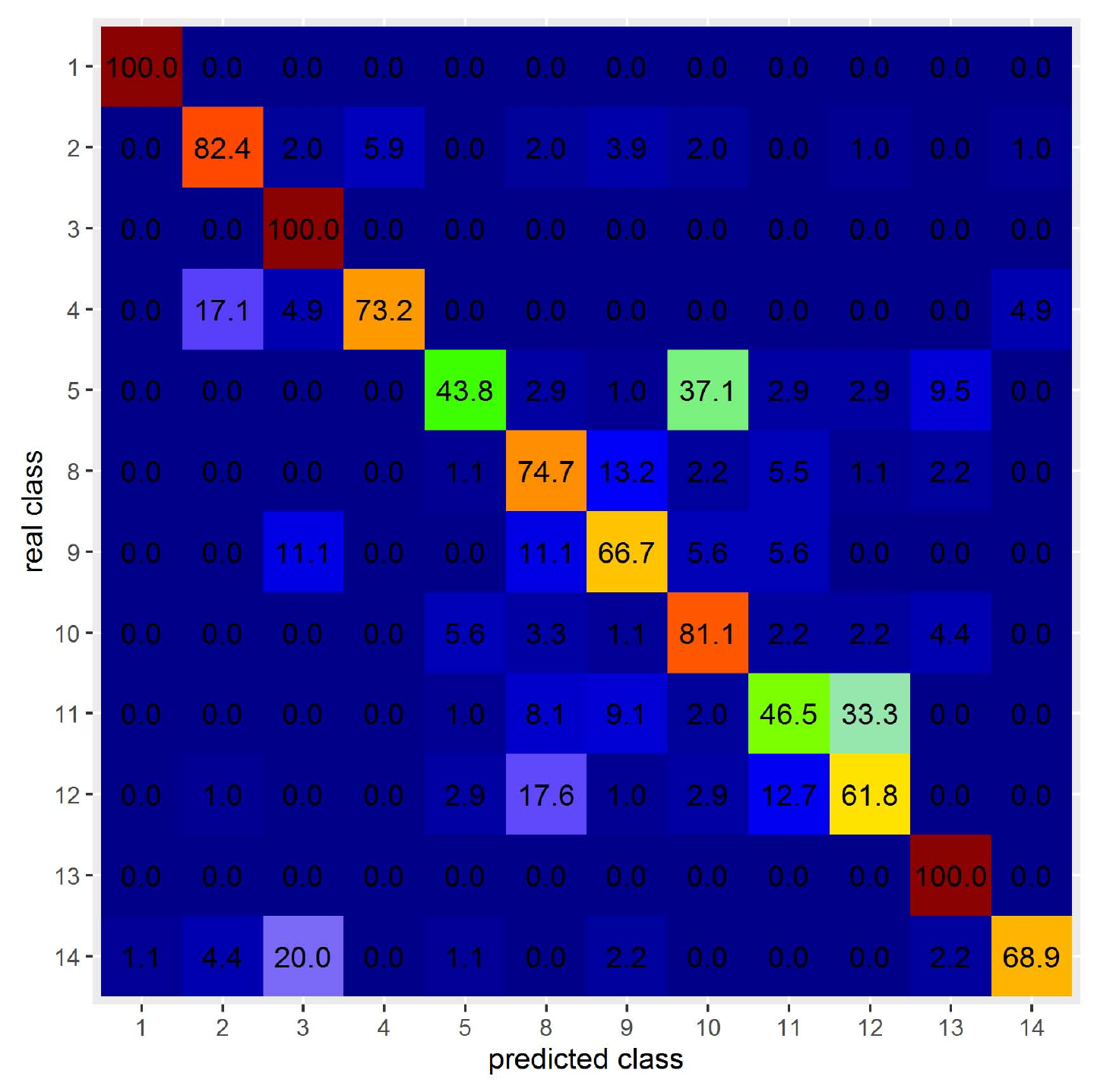} &
            \includegraphics[width=0.24\textwidth]{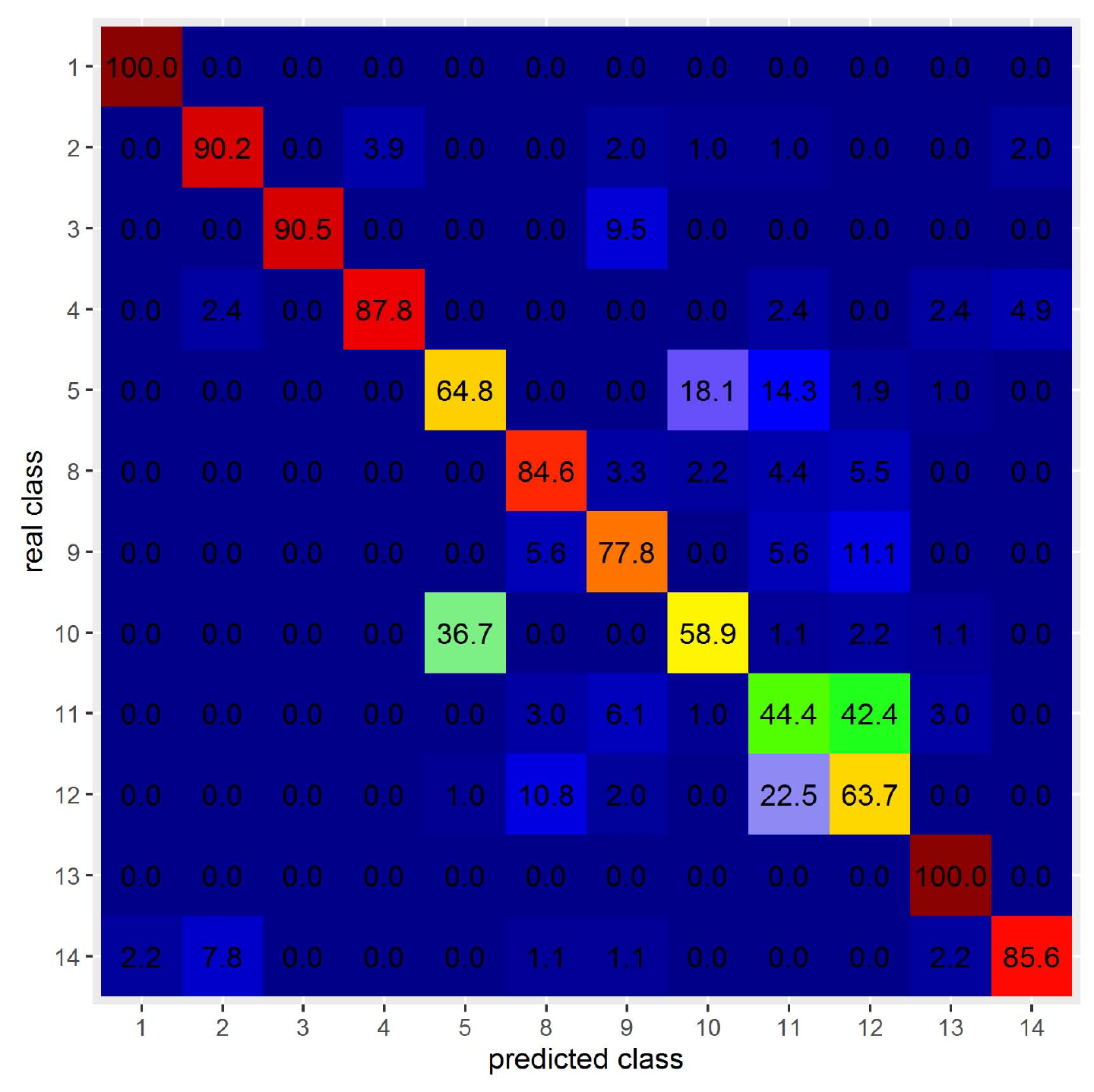} \vspace{0.5cm} \\ 
         \end{tabular}  
                
         RP-RGB $BoVW^{1000}_{RGBSIFT}$ \\
         \includegraphics[width=0.24\textwidth]{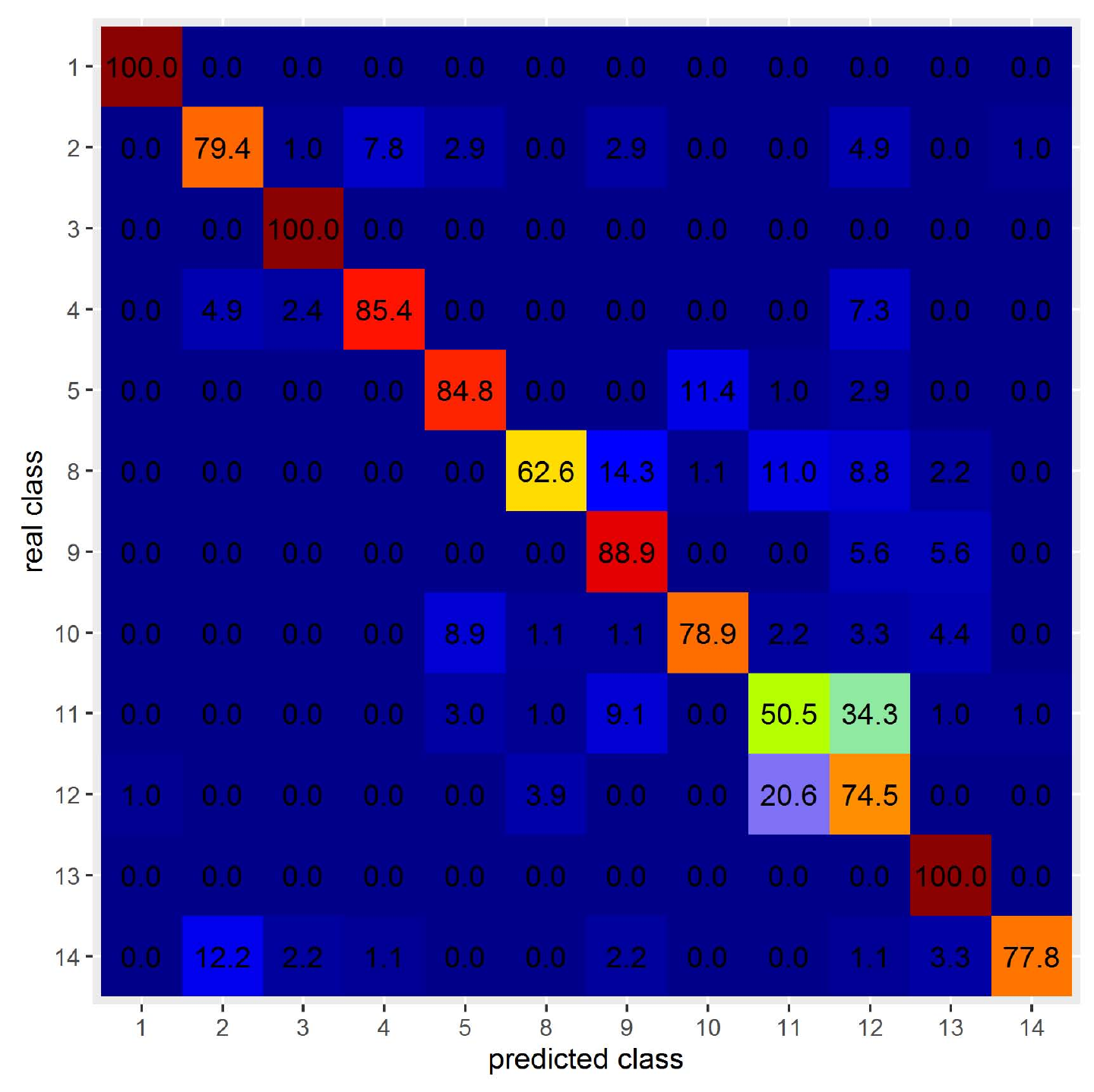}
		 \includegraphics[height=4.2cm]{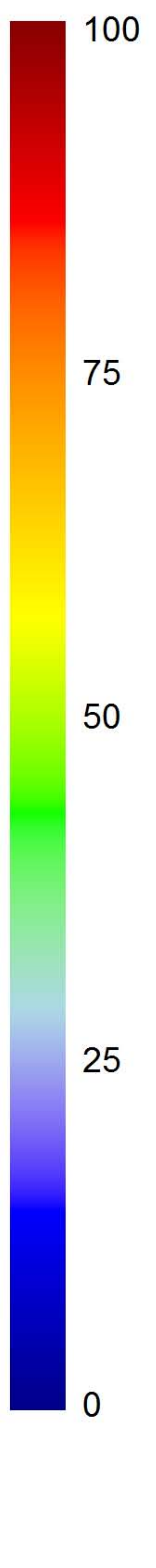} \vspace{-0.3cm}
  
    \end{center}
    \caption{Confusion matrices of the best performing methods of each feature modality. We can note that activities 11 and 12 (sit down chair and stand up chair) are among the most difficult to discriminate. The class numbers in each axis refer to the numbers shown in Table~\ref{tab_wharf}.}
    \label{fig_conf_matrix}
  \end{figure}

To illustrate the results per class, we show in Figure~\ref{fig_conf_matrix} the confusion matrices of the best performing methods of each modality: time-domain feature (quantile), frequency-domain feature (dftbands), RP-gray $BoVW^{1000}_{SIFT}$, RP-gray-concat $BoVW^{1000}_{SIFT}$, and RP-RGB $BoVW^{1000}_{RGBSIFT}$.

The most confusing classes for most of the methods are: 5 and 10 (drink glass and pour water), 2 and 14 (climb stairs and walk), and 11 and 12 (sit down chair and stand up chair).
In the case of drink glass and pour water, the movements can be very similar in terms of accelerometer data.
Climb stairs and walk, and sit down and stand up chair are also very similar.
Classes 1, 3, 13 (brush teeth, comb hair, use telephone) are the easiest ones. 
However, they are among the classes with less samples, so only few samples were available for testing.

For the features extracted directly from sensor data (quantile and dftbands), the most confusing classes are 8 and 9 (get up bed and lie down bed) and 11 and 12 (sit down chair and stand up chair).
Classes 8 and 9, however, can be better distinguished by some methods based on RPs.
For the best method (RP-RGB $BoVW^{1000}_{RGBSIFT}$), the most confusing classes are 11 and 12 (sit down chair and stand up chair).
With this per-class and per-method analysis, we can also identify opportunities for fusing the different approaches.

In general, by considering only accelerometer data, some activities will be difficult to distinguish.
In order to obtain improvements in accuracy, other sensors may be necessary, like gyroscopes.

In summary, we can point that the use of visual features over RGB Recurrence Plots, specially BoVW based on RGB SIFT and using max pooling with spatial pyramids (maxSPM), is a promising direction for the recognition of human activities based on inertial sensor data.


\section{Conclusions}
\label{conclusions}
We presented an approach for human activity recognition based on inertial sensors by employing recurrence plots (RP) and visual descriptors.
As RPs generate texture visual patterns, the proposed approach transforms the problem of sensor data classification into a problem of texture classification.
The proposed pipeline is the following: compute RPs from sensor data (time series), compute visual features from RPs and use them in a machine learning protocol.
Results for classifying human activities based on accelerometer data showed that the proposed approach obtained the highest accuracy rates in comparison with traditional statistical time- and frequency-domain features.
The use of bags of visual words descriptors over RGB plots, in which each RGB channel corresponds to the RP of an independent accelerometer axis, presented the best results.
The use of spatial pyramids was also important for distinguishing similar plots with different spatial arrangement of textures.

As future work, we would like to explore other datasets including other sensor data, like gyroscopes and barometers, as well as datasets including other human activities.
Considering the visual descriptors, we would like to perform a correlation analysis of the visual descriptors for finding possibilities for fusing them and improving the results.
In addition, an option to improve features from convolutional neural networks (CNN) is to perform fine tuning of existing CNNs using the RP data or even to design a specific network for texture recognition. 


\bibliographystyle{model2-names}
\bibliography{refs}

%

\end{document}